\documentclass[authoryear,preprint,a4paper]{elsarticle}

\usepackage{ifpdf}

\usepackage{graphicx}
\usepackage{epstopdf}
\usepackage{booktabs}
\usepackage{multirow}
\usepackage{float}
\usepackage{amsmath}
\usepackage{array}
\usepackage[justification=centering]{caption}
\usepackage{fixltx2e}
\usepackage{stfloats}
\usepackage{xcolor}
\usepackage{amssymb}
\usepackage{amsthm}
\usepackage{url}

\usepackage[ruled,vlined,linesnumbered]{algorithm2e}

\usepackage{longtable}

\usepackage{subfigure}

\usepackage{ifpdf}
\usepackage{graphicx}
\usepackage{float}
\usepackage{epstopdf}
\usepackage{amsmath}
\usepackage{array}
\usepackage[justification=centering]{caption}
\usepackage{subfigure}
\usepackage{stfloats}
\usepackage{amssymb}
\usepackage{amsthm}
\usepackage{url}
\usepackage{multirow,booktabs,color,soul,threeparttable}
\usepackage{rotating}

\usepackage{makecell}
\usepackage{multirow}
\usepackage[T1]{fontenc}
\usepackage{xcolor}
\usepackage{colortbl}

\usepackage[colorlinks=true, linkcolor=blue, citecolor=green, urlcolor=red]{hyperref} 

\definecolor{hl}{rgb}{0.75,0.75,0.75}
\sethlcolor{hl}

\makeatletter
\newcommand{\removelatexerror}{\let\@latex@error\@gobble}
\makeatother
\graphicspath{{Figs/}}
\usepackage{lineno} 

\journal{Elsevier}

\begin{document}
	
	\begin{frontmatter}
		
		\title{Composite Indicator-Guided Infilling Sampling for Expensive Multi-Objective Optimization}
		
		\author[cug]{Huixiang Zhen}
		\ead{zhenhuixiang@cug.edu.cn}

        \author[cug]{Xiaotong Li}
		\ead{Lxt@cug.edu.cn}
		
		\author[cug]{Wenyin Gong\corref{cor1}}
		\cortext[cor1]{Corresponding author}
		\ead{wygong@cug.edu.cn}

        \author[cug_SGG]{Xiangyun Hu}
		\ead{xyhu@cug.edu.cn}
		
		\address[cug]{School of Computer Science, China University of Geosciences, Wuhan 430074, China.}

        \address[cug_SGG]{School of Geophysics and Geomatics, China University of Geosciences, Wuhan 430074, China.}
		
		\begin{abstract}
            In expensive multi-objective optimization, where the evaluation budget is strictly limited, selecting promising candidate solutions for expensive fitness evaluations is critical for accelerating convergence and improving algorithmic performance. However, designing an optimization strategy that effectively balances convergence, diversity, and distribution remains a challenge. To tackle this issue, we propose a composite indicator-based evolutionary algorithm (CI-EMO) for expensive multi-objective optimization. In each generation of the optimization process, CI-EMO first employs NSGA-III to explore the solution space based on fitness values predicted by surrogate models, generating a candidate population. Subsequently, we design a novel composite performance indicator to guide the selection of candidates for real fitness evaluation. This indicator simultaneously considers convergence, diversity, and distribution to improve the efficiency of identifying promising candidate solutions, which significantly improves algorithm performance. The composite indicator-based candidate selection strategy is easy to achieve and computes efficiency. Component analysis experiments confirm the effectiveness of each element in the composite performance indicator. Comparative experiments on three benchmark test sets and real-world problems demonstrate that the proposed algorithm outperforms five state-of-the-art expensive multi-objective optimization algorithms.
		\end{abstract}
		
		
	\end{frontmatter}
	
	\nolinenumbers
	
	\section{Introduction}\label{Sec_I}
        \par In recent decades, evolutionary algorithms (EAs) have demonstrated remarkable effectiveness in addressing multi- and many-objective optimization problems~\citep{Cardona2020-indicator, Zhang2024-enhancing, Rostamian2024-survey}. These population-based metaheuristic approaches are highly effective at generating diverse and well-converged solutions within a single optimization run~\citep{Liu2023-survey, Zhang2024-paretotracker}. Optimization problems with two or three objectives are typically referred to as multi-objective optimization, while those involving more than three objectives are categorized as many-objective optimization~\citep{Li2015-survey}. A wide variety of multi- and many-objective optimization algorithms have been introduced in the literature, which are commonly grouped into three categories: dominance-based~\citep{Zhou2024-dominance}, decomposition-based~\citep{Li2024-Decomposition}, and indicator-based approaches~\citep{Cardona2020-indicator}.

        \par However, in real-world engineering applications, the objective evaluations of many optimization problems often lack explicit mathematical expressions. Instead, these evaluations typically rely on software simulations or physical experiments, both of which can be highly time-consuming~\citep{Zhen2021-TS-DDEO, ZHAI2024121374}, such as airfoil design~\citep{Dai2024-multi}, lightweight design of vehicle~\citep{Li2024-integrated}, and network architecture search~\citep{Li2025-automatic}. These kinds of problems are named expensive multi/many-objective optimization problems (EMOPs/EMaOPs). In these problems, objective function evaluation generally is computationally time-consuming, which hinders the application of traditional multi-objective evolutionary algorithms (MOEAs)~\citep{Li2022-survey}. For this reason, surrogate models are widely employed in EMOPs to reduce computational costs by using efficient machine learning methods to approximate the real objective values~\citep{Wei2024-AIEA, Zhang2024-DISK}. Surrogate-assisted evolutionary algorithms (SAEAs) leverage previously accumulated data to build surrogate models. During the optimization process, candidate solutions are initially evaluated using these trained models, and only the most promising ones undergo expensive real fitness evaluations (FEs). This approach effectively reduces the number of expensive evaluations, thereby saving computational time~\citep{Jin2018-survey, He2023-survey}.

        \begin{figure}[ht]
    		\centering
    		\resizebox{\linewidth}{!}{
    			\includegraphics[scale=1]{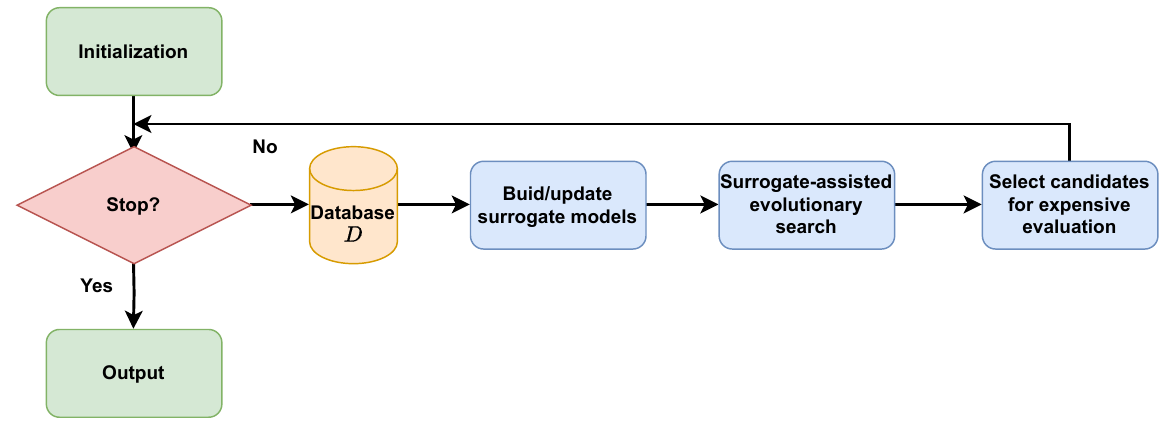}
    		}
    		\caption{Flowchart of a general surrogate-assisted MOEAs.}
    		\label{Figure_1_MOEAs}
	\end{figure}

        \par Fig.~\ref{Figure_1_MOEAs} shows the flowchart of general surrogate-assisted MOEAs. In solving EMOPs, there are usually only a few hundred expensive FEs available~\citep{Zhang2024-DISK, Yang2024-HES-EA}. Therefore, all data evaluated by the real fitness function are generally archived in a database. After initializing the database, optimization consists of three main parts:
        \begin{enumerate}
        \item \textbf{Build/update surrogate models:} Surrogate models are built based on data in the database. The common surrogate models include Gaussian Process model (GP, also named Kriging model)~\citep{Schulz2018-tutorial, Wang2025-kriging}, radial basis function network (RBFN)~\citep{CHEN2023120826}, and support vector machine (SVM)~\citep{Sonoda2022-multiple} et al. During the optimization process, the surrogate models are updated in response to changes in the database.
        \item \textbf{Surrogate-assisted evolutionary search:} A MOEA is employed to perform an evolutionary search across the approximate fitness landscapes constructed by the surrogate models over multiple generations. Finally, a candidate population is output for further selection. Ultimately, a candidate population $\mathcal{P^*}$ is produced for further selection.
        \item \textbf{Select candidates for expensive evaluation:}
        To save the number of real evaluations, a subset of the most promising candidates, denoted as $\mathcal{C}$, is selected from $\mathcal{P^*}$. These candidates are then evaluated based on the original expensive objectives and then stored in $\mathcal{D}$.
        \end{enumerate}
        Repeat the above steps until all expensive evaluations are exhausted, and finally output all non-dominated solutions in the database.

        \par By instantiating specific parts of the generic framework, numerous surrogate-assisted MOEAs have been developed to address expensive multi-objective optimization problems~\citep{Zhang2024-DISK, Liu2023-solving, Wei2024-AIEA}. Since the real fitness evaluation budget is very limited, it is more advantageous to allocate fitness evaluations to the most promising candidate solutions. This raises a key challenge: how to efficiently identify promising solutions during evolutionary search and candidate selection.

        \par For this issue, some representative works have been presented recently. On one hand, some multiobjective efficient global optimization (MOEGO) methods have been proposed based on Bayesian optimization, in which an appropriate expected improvement (EI) function is designed to sample candidates for expensive evaluation~\citep{Xu2025-adaptive, Zhan2020-expected}. Zhao et al. present an expected direction-based hypervolume improvement to select candidates. After that, an improved method selects a subset of query points based on the lower bound of R2-based EI from the candidates which are generated by the Tchebycheff decomposition paradigm~\citep{Zhao2024-R2DEGO}. In these ways, some solutions with uncertainty increments can be retained to improve the accuracy of the alternative model. However, due to the limited number of training samples available for surrogate models, the estimation of uncertainty information can often be inaccurate, which adversely impacts the ability to preserve promising solutions~\citep{Zhang2024-DISK}. On the other hand, Chugh et al. proposed selecting one individual from each cluster after the Kriging-assisted reference vector-guided EA search, based on either the angle penalized distance or maximum uncertainty~\citep{Chugh2016-KRVEA}. Song et al. divide optimization into three states after generating candidates using Two\_Arch2~\citep{Wang2014-Two_arch2}, and use their respective sampling strategies (convergence, diversity, and uncertainty sampling) to select candidate solutions, respectively~\citep{Song2021-KTA2}. In addition to methods that use sampling strategies for a single purpose, Qin et al. proposed a performance indicator that considers both convergence and diversity properties of candidate solutions by two types of distance calculation~\citep{Qin2023-EMMOEA}. However, current approaches lack consideration of the distribution of solutions. Overall, the above candidate solution selection methods still do not balance convergence, diversity, and distribution in the search process. Distribution means the uniform distribution of the solution set on the Pareto front. To differentiate from the distribution, in this paper, the term "diversity" emphasizes the spread of solution, that is, the expansion of the solution boundary, and whether the extreme areas of the target space are fully explored~\citep{Cardona2020-indicator}.
        
        \par To develop an optimization strategy that effectively balances convergence, diversity, and distribution, while simplifying algorithm design via a modular approach, this paper proposes a novel composite indicator infilling sampling-based expensive multi-objective optimization (CI-EMO) method. The main contributions of this research are as follows:

        \begin{enumerate}
        \item This paper introduces a new composite indicator, which integrates three weighted performance metrics of candidate solutions: convergence, diversity, and distribution. This indicator is used to select the most promising candidate solution for real fitness evaluation from the generated population. The composite indicator is achieved based on the objective space position of the candidate solutions, which is computing efficiency, and easy to achieve and usage.
        \item Experiments validate that the proposed composite indicator is better than single one. The composite indicator-guided evolutionary algorithm is a promising direction for candidate selection in SAEAs. 
        \item The composite indicator is incorporated into surrogate-assisted MOEAs for expensive multi-objective optimization. At each iteration, CI-EMO first employs GP-assisted NSGA-III~\citep{Deb2013-NSGA-III} as the optimizer to generate the candidate population. The composite indicator is then used to select the most promising candidate for real fitness evaluation, achieving efficient sampling and improving the solution set's distribution. 
        
        \end{enumerate}
        \par The rest parts of this paper are organized as follows. Section~\ref{Sec_II} introduces the background of this work. In Section~\ref{Sec_III}, we introduce our proposed composite indicator and CI-EMO algorithm. In Section~\ref{Sec_IV}, a series of experiments are designed to validate the performance of our proposed method. Finally, Section~\ref{Sec_V} offers a conclusion.

        \section{BACKGROUND}\label{Sec_II}
        \subsection{Problem Formulation}\label{Sec_II_A}
        \par An EMOP can be described as:
        $$
        \begin{array}{ll}
        \text{minimize} & F(\boldsymbol{x}) = \left(f_1(\boldsymbol{x}), \dots, f_m(\boldsymbol{x})\right)^T \\
        \text{subject to} & \boldsymbol{x} \in \Omega \subseteq \mathbb{R}^d
        \end{array}
        $$
        where $\boldsymbol{x}$ represents the decision variables, and $\Omega \subseteq \mathbb{R}^d$ defines the feasible decision space. The function $F: \Omega \to \mathbb{R}^m$ is a vector function containing $m$ continuous objectives. In the context of this study, we focus on cases where evaluating the functions is computationally expensive but can be done in parallel, for example, using multiple processors or machines~\citep{Liu2023-survey}.

        \par Let $\boldsymbol{x}_a$ and $\boldsymbol{x}_b$ represent two vectors in the decision space. Vector $\boldsymbol{x}_a$ is said to dominate $\boldsymbol{x}_b$ (denoted as $\boldsymbol{x}_a \prec \boldsymbol{x}_b$) if and only if $f_i(\boldsymbol{x}_a) \leq f_i(\boldsymbol{x}_b)$ holds for all $i = 1, \dots, m$, with at least one objective index $j$ satisfying $f_j(\boldsymbol{x}_a) < f_j(\boldsymbol{x}_b)$. A solution $\boldsymbol{x}^* \in \Omega$ is deemed Pareto optimal if no other solution dominates it. The corresponding objective vector, $F(\boldsymbol{x}^*)$, is referred to as the Pareto optimal objective vector. The collection of all Pareto optimal solutions constitutes the Pareto set (PS), while the set of all associated Pareto optimal objective vectors defines the Pareto front (PF)~\citep{Yu1974-cone}.

        \subsection{Gaussian Process Model}\label{Sec_II_A}
        \par Gaussian Process model is a machine learning technique used for spatial interpolation, offering predictions of objective values for solutions along with an estimate of the uncertainty in these predictions~\citep{Schulz2018-tutorial}.

        \par For building and training the Gaussian Process model, we utilize the DACE toolbox in Matlab. The GP model assumes a Gaussian process prior distribution over the function space, expressed as:
        $$
        f(\mathbf{x}) \sim \mathcal{GP}\left( \mu(\mathbf{x}), \text{Cov}(\mathbf{x}, \mathbf{x'}; \boldsymbol{\theta}) \right),
        $$
        where $\mu(\mathbf{x})$ represents the mean function (often simplified to zero for convenience), and $\text{Cov}(\mathbf{x}, \mathbf{x'}; \boldsymbol{\theta})$ denotes the covariance function, which depends on the parameters $\boldsymbol{\theta}$ and describes the relationship between outputs $f(\mathbf{x})$ and $f(\mathbf{x'})$. For two points $\mathbf{x}_i$ and $\mathbf{x}_j$ in the decision space, the covariance function is defined as:
        $$
        \text{Cov}(\mathbf{x}_i, \mathbf{x}_j; \boldsymbol{\theta}) = \exp \left(-\sum_{k=1}^d \theta_k (x_{i,k} - x_{j,k})^2 \right),
        $$
        where $\boldsymbol{\theta} = [\theta_1, \theta_2, \dots, \theta_d]$ is the vector of hyperparameters that are estimated by maximizing the likelihood of $n$ observations from the dataset $\mathbb{D}$, as given by:
        $$
        \begin{aligned}
        \max_{\boldsymbol{\theta}} \left( -\frac{\ln \hat{\sigma}^2 + \ln \|\mathbf{R}\|}{2} \right), \\
        \hat{\mu} = \frac{\mathbf{1}^T \mathbf{R}^{-1} \mathbf{y}}{\mathbf{1}^T \mathbf{R}^{-1} \mathbf{1}}, \\
        \hat{\sigma}^2 = \frac{(\mathbf{y} - \mathbf{1} \hat{\mu})^T \mathbf{R}^{-1} (\mathbf{y} - \mathbf{1} \hat{\mu})}{n}.
        \end{aligned}
        $$
        where, $\mathbf{1}$ represents an all-ones vector, $\mathbf{y}$ means objective function value vector and the matrix $\mathbf{R}$ is the covariance matrix, defined as:
        $$
        \mathbf{R} = \begin{bmatrix}
        \text{Cov}(\mathbf{x}_1, \mathbf{x}_1; \boldsymbol{\theta}) & \cdots & \text{Cov}(\mathbf{x}_1, \mathbf{x}_n; \boldsymbol{\theta}) \\
        \vdots & \ddots & \vdots \\
        \text{Cov}(\mathbf{x}_n, \mathbf{x}_1; \boldsymbol{\theta}) & \cdots & \text{Cov}(\mathbf{x}_n, \mathbf{x}_n; \boldsymbol{\theta})
        \end{bmatrix}.
        $$

        \par To predict the objective value at a new input point $\mathbf{x}$, we compute:
        $$
        \hat{y}(\mathbf{x}) = \hat{\mu} + \mathbf{r}^T \mathbf{R}^{-1} (\mathbf{y} - \mathbf{1} \hat{\mu}),
        $$
        where $\mathbf{r}$ is the vector of covariances between the new point $\mathbf{x}$ and each training point. The uncertainty in this prediction is given by:
        $$
        \hat{s}^2(\mathbf{x}) = \hat{\sigma}^2 \left[ 1 - \mathbf{r}^T \mathbf{R}^{-1} \mathbf{r} + \frac{\left( 1 - \mathbf{1}^T \mathbf{R}^{-1} \mathbf{r} \right)^2}{\mathbf{1}^T \mathbf{R}^{-1} \mathbf{1}} \right],
        $$
        where $\mathbf{r}$ is the row vector corresponding to the point $\mathbf{x}$ in the covariance matrix $\mathbf{R}$.

        \section{PROPOSED ALGORITHM}\label{Sec_III}
        \subsection{CI-EMO Framework}\label{Sec_III_A}
        \par
\begin{figure}[ht]
    		\centering
    		\resizebox{\linewidth}{!}{
    			\includegraphics[scale=1]{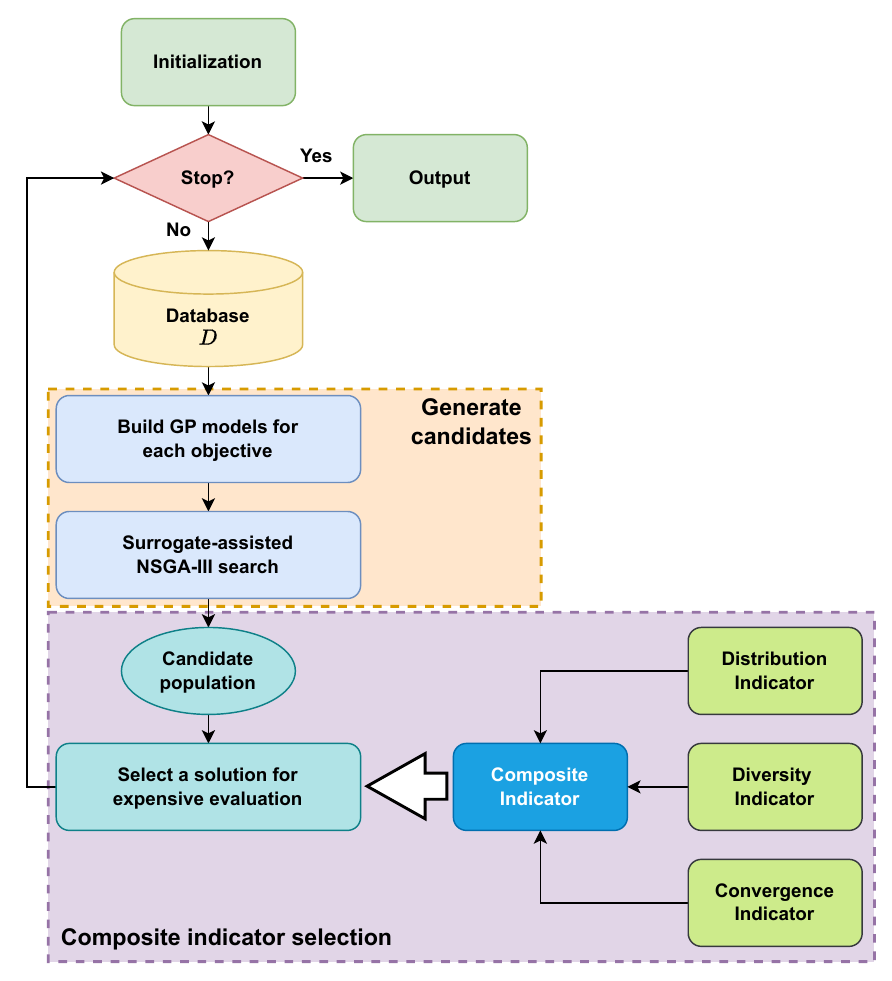}
    		}
    		\caption{The framework of CI-EMO.}
    		\label{Figure_2_framework}
	\end{figure}

 The framework of CI-EMO is illustrated in Fig.~\ref{Figure_2_framework}. The CI-EMO algorithm begins by initializing a database through Latin Hypercube Sampling (LHS)~\citep{Iman2008-latin}. Each generation of the algorithm consists of two key stages: (1) Candidates generation and (2) Composite indicator-based selection. During the optimization process, CI-EMO first constructs Gaussian Process surrogate models for each objective based on the database. It then leverages the surrogate-assisted NSGA-III~\citep{Deb2013-NSGA-III} (SA-NSGA-III) algorithm to explore the search space over a predefined number of generations, resulting in a candidate population. Subsequently, CI-EMO employs a composite indicator that synthesizes three distinct metrics, i.e., distribution, diversity, and convergence, to select the most promising sample point from the candidate population for expensive fitness evaluation. The evaluated data point is then incorporated into the database, facilitating the refinement of surrogate models for subsequent generations.

        \par The pseudocode of CI-EMO is presented in Algorithm~\ref{alg:CI-EMO}. Initially, the number of initial samples, $\mathcal{N}_0$, and the maximum allowable expensive function evaluations, $\mathcal{N}_{total}$, are specified. CI-EMO employs LHS to generate $\mathcal{N}_0$ samples, which are then evaluated using expensive fitness function. These evaluated samples form a database, denoted as $\mathcal{D}$. During the optimization process, CI-EMO constructs $m$ surrogate models, $\mathcal{M} = \{{M_i}\}_i^{m}$, where each model corresponds to one objective. Subsequently, the surrogate-assisted NSGA-III (SA-NSGA-III) algorithm is applied to explore the search space and generate a candidate population, $P^*$. Next, the composite indicator values of $P^*$ are computed, and a query point $\boldsymbol{x}$ with the highest composite indicator value is selected. The algorithm then evaluates the objective vectors, $F\left(\boldsymbol{x}\right)$, of the selected point. The new sample point is incorporated into the database. Once the expensive fitness evaluation times are exhausted, the algorithm outputs the non-dominated solutions in the database.

        \begin{algorithm}[!htbp]
    	\caption{\textbf{CI-EMO}}
    	\label{alg:CI-EMO}
    	\KwIn{$\mathcal{N}_0$: The number of initial samples; $\mathcal{N}_{total}$: The total number of expensive function evaluations;}
    		\KwOut{Nondominated solutions of all evaluated solutions in database $\mathcal{D}$;}
            \textbf{Initialize:} Generate $\mathcal{N}_0$ samples $\left\{\boldsymbol{x}^i\right\}_{i=1}^{\mathcal{N}_0}$ from $\Omega$ via Latin Hypercube Sampling and observe the objective vectors $F\left(\boldsymbol{x}^i\right)$ for each sample\;
            Database $\mathcal{D} \leftarrow\left\{\left(\boldsymbol{x}^i, F\left(\boldsymbol{x}^i\right)\right)\right\}_{i=1}^{\mathcal{N}_0}$\;
            \While{ $|\mathcal{D}|<\mathcal{N}_{\text {total}}$ }{
                Fit $m$ surrogate models $\mathcal{M} = \{{M_i}\}_i^{m}$, one for each objective\;
                $P^* \leftarrow$ \textbf{SA-NSGA-III}($\mathcal{D}$, $\mathcal{M}$)\;
                Choose a query point $\boldsymbol{x}$ from $P^*$ by \textbf{CI}($P^*$, $\mathcal{D}$)\;
                $\text{Observe the objective vectors } F\left(\boldsymbol{x}\right)$\;
                $\mathcal{D} \leftarrow \mathcal{D} \cup\left\{\boldsymbol{x}, F\left(\boldsymbol{x}\right)\right\}$\;
            }
            \textbf{Return}: All nondominated solutions in $\mathcal{D}$\;
	\end{algorithm}

        \subsection{Generate candidates}\label{Sec_III_A}
        \par  To enhance the convergence rate of the algorithm and effectively identify promising sample points, CI-EMO leverages SA-NSGA-III during the optimization process to explore the solution space and generate a candidate population for subsequent selection. The pseudocode of SA-NSGA-III is presented in Algorithm 2. Initially, the input consists of the database $\mathcal{D}$ and surrogate models $\mathcal{M}$. The algorithm generates a set of reference vectors, denoted as $V$, and computes the ideal point $Z$ based on the database. Using the environmental selection mechanism of NSGA-III~\citep{Deb2013-NSGA-III}, a non-dominated population $P$ is obtained. The subsequent search procedure follows the NSGA-III framework, where the fitness values of offspring are predicted by surrogate models. Specifically, offspring $O$ are generated through crossover and mutation operations, and then merged with population $P$. Once the maximum number of iterations, $T_{max}$, is reached, the algorithm outputs the candidate population $P^*$.

        \begin{algorithm}[!htbp]
    	\caption{\textbf{SA-NSGA-III} ($\mathcal{D}$, $\mathcal{M}$)}
    	\label{alg:SA-NSGA-III}
    	\KwIn{Database $\mathcal{D}$; Surrogate models $\mathcal{M} = \{{M_i}\}_i^{m}$;}
            \KwOut{A candidate population $P^*$;}
            $V \leftarrow $ Generate a set of reference vectors\;
            $Z \leftarrow $ Get the minimum value corresponding to each objective in the $\mathcal{D}$\;
            $P \leftarrow $ Environmental selection of NSGA-III($\mathcal{D}$, $Z$, $V$)\;
            $t = 1$\;
            \While{ $t<T_{max}$ }{
                $O \leftarrow$ Performing crossover and mutation on $P$\;
                $P \leftarrow O \bigcup P $\;
                Evaluation by $\mathcal{M}$\;
                $P \leftarrow $ Environmental selection of NSGA-III($P$, $Z$, $V$)\;
                $t = t + 1$\;
            }
            $P^* \leftarrow P$\;
	\end{algorithm}

        \begin{algorithm}[!htbp]
    	\caption{$\textbf{CI}$ ($P^*$, $\mathcal{D}$)}
    	\label{alg:CI}
    	\KwIn{Candidate population $P^*$ (which consists of decision variables $\boldsymbol{x}^*$ and its predicted fitness $\hat{\boldsymbol{F}}$); Database $\mathcal{D}$;}
            \KwOut{The individual $\boldsymbol{x}$ selected for expensive evaluation;}
            Performs non-dominated sorting on $\mathcal{D}$ to obtain the non-dominated solution set $\mathcal{D}_{nd}$\;
            Distribution indicator $\mathbf{I_1}$ is calculated by Eqs.~\ref{eq:angle1}-\ref{eq:I1}\;
            Normalize objective values of all solutions in $P^*$ and $\mathcal{D}$ by Eqs.~\ref{eq:normalization1}-\ref{eq:normalization2}\;
            Diversity indicator $\mathbf{I_2}$ is calculated by Eqs.~\ref{eq:crowdedness}-\ref{eq:I2}\;
            Convergence indicator $\mathbf{I_3}$ is calculated by Eqs.~\ref{eq:convergence}-\ref{eq:I3}\;
            Generate random variables $r_1, r_2, r_3$ between 0 and 1\;
            Composite indicator $\mathbf{CI} = r_1\cdot\mathbf{I_1} + r_2\cdot\mathbf{I_2} + r_3\cdot\mathbf{I_3}$\;
            $\boldsymbol{x} \leftarrow $ Select the solution with maximum $\mathbf{CI}$ value in $P^*$\;
	\end{algorithm}

        \subsection{Composite indicator-based selection}\label{Sec_III_A}
        \par The combination of indicators is an important method to generate new selection mechanisms with different requirements~\citep{Cardona2020-indicator, Falcon2022-construction}. To improve the sampling effect for the expensive multi-objective problem, CI-EMO combines three sampling metrics into a composite indicator. The three metrics consider distribution, diversity, and convergence, respectively. Algorithm~\ref{alg:CI} presents the pseudocode for the composite indicator selection used in CI-EMO. The composite performance of each candidate solution in the $P^*$ will be evaluated in terms of distribution, diversity, and convergence metrics. Note that, the three component indicators are calculated only based on $P^*$ and $\mathcal{D}$. As shown in algorithm~\ref{alg:CI}, before calculating the distribution indicator, the algorithm performs non-dominated sorting on $\mathcal{D}$ then obtains the non-dominated solution set $\mathcal{D}_{nd}$. The distribution indicator is calculated based on the original objective value. However, before calculating diversity and convergence indicators, the normalized objective function of solutions $P^*$ and $\mathcal{D}$ is calculated, respectively, as follows:
        \begin{equation}
            \label{eq:normalization1}
            \tilde{\hat{\boldsymbol{F}}}(\boldsymbol{x})=\frac{\hat{\boldsymbol{F}}(\boldsymbol{x})-\boldsymbol{F}_{\min}}{\boldsymbol{F}_{\max}-\boldsymbol{F}_{\min}},
            \quad \boldsymbol{x} \in P^*
        \end{equation}
        \begin{equation}
            \label{eq:normalization2}
            \tilde{\boldsymbol{F}}(\boldsymbol{x})=\frac{\boldsymbol{F}(\boldsymbol{x})-\boldsymbol{F}_{\min}}{\boldsymbol{F}_{\max}-\boldsymbol{F}_{\min}},
            \quad \boldsymbol{x} \in \mathcal{D}
        \end{equation}
        where $\boldsymbol{F}_{\min}$ and $\boldsymbol{F}_{\max}$ represent the minimum and maximum values, respectively, for each objective in the database $\mathcal{D}$. Subsequently, the diversity and convergence indicators are computed, and the three indicators are integrated using random weights to obtain the composite indicator, CI. Finally, the solution with the maximum CI value in $P^*$ is selected as the final candidate solution, $\boldsymbol{x}$, and output. Note that random weight setting can enhance the robustness of the algorithm, which will be verified in the experiment section. The three component indicators of CI are described in detail as follows:

        \begin{enumerate}
            \item \textbf{Distribution indicator ($\mathbf{I_{1}}$)}: The distribution of a candidate solution $x$ is determined by measuring the minimum angle, $\theta(\boldsymbol{x})$, between the solution and the set $\mathcal{D}_{nd}$. It is worth noting that both the distribution indicator and the diversity indicator increase the diversity of samples, but compared with the diversity indicator, the distribution indicator here emphasizes the uniform distribution of the distributed solution set on the Pareto front. The value of $\theta(\boldsymbol{x})$ for each solution in $P^*$ is computed using Eqs.~\ref{eq:angle1} and \ref{eq:angle2}. 
            \begin{equation}
                \label{eq:angle1}
                \theta_{\boldsymbol{x} \boldsymbol{y}}=\arccos \frac{\sum_{i=1}^m\left(\hat{f}_i(\boldsymbol{x}) \cdot f_i(\boldsymbol{y})\right)}{\sqrt{\sum_{i=1}^m \hat{f}_i(\boldsymbol{x})^2} \cdot \sqrt{\sum_{i=1}^m f_i(\boldsymbol{y})^2}},
            \end{equation}
            \begin{equation}
                \label{eq:angle2}
                \theta(\boldsymbol{x}) =
                \min _{\boldsymbol{y} \in \mathcal{D}_{nd}}
                \theta_{\boldsymbol{x} \boldsymbol{y}},
                \quad \boldsymbol{x} \in P^*.
            \end{equation}
            Finally, $\theta(\boldsymbol{x})$ is normalized to obtain $\mathbf{I}_1$. 
            \begin{equation}
                \label{eq:I1}
                \mathbf{I}_1 = \mathrm{Normalize}\left( \theta(\boldsymbol{x}) \right),
                \quad \boldsymbol{x} \in P^*.
            \end{equation}
            Normalization is the process of scaling data to a specific range to ensure uniformity and comparability. Here, we use min-max normalization to the range [0, 1], as shown in Eq.~\ref{eq:normalization}, where $\theta(\boldsymbol{x})_{\min}$ and $\theta(\boldsymbol{x})_{\max}$ are the minimum and maximum values in all $\theta(\boldsymbol{x})$.
            \begin{equation}
                \label{eq:normalization}
                \mathrm{Normalize}\left( \theta(\boldsymbol{x}) \right)=\frac{\theta(\boldsymbol{x})-\theta_{\min}}{\theta_{\max}-\theta_{\min}}, \boldsymbol{x} \in P^*
            \end{equation}

            \item \textbf{Diversity indicator ($\mathbf{I_{2}}$)}: The diversity of a candidate solution $\boldsymbol{x}$ is defined based on the crowdedness surrounding the solution. It is also termed as MaxMin diversity indicator~\citep{Pereverdieva2025-comparative}. Compared to distribution, the indicator emphasizes the spread of solution, that is, the expansion of the solution boundary, and whether the extreme areas of the target space are fully explored~\citep{Cardona2020-indicator}. It is quantified by computing the distance in the objective space to its nearest neighbor $\tilde{{\boldsymbol{F}}}\left(\boldsymbol{y}\right)$, as expressed by the Eq.~\ref{eq:crowdedness}. $\mathbf{I}_2$ is calculated by normalizing $d_{c}$. 
            \begin{equation}
                \label{eq:crowdedness}
                d_{c}\left(\boldsymbol{x}\right)=\min _{\boldsymbol{y} \in \mathcal{D}} \left\|\tilde{{\hat{\boldsymbol{F}}}}\left(\boldsymbol{x}\right)-\tilde{{\boldsymbol{F}}}\left(\boldsymbol{y}\right)\right\|,
                \quad \boldsymbol{x} \in P^*.
            \end{equation}
            \begin{equation}
                \label{eq:I2}
                \mathbf{I}_2 = \mathrm{Normalize}\left( d_{c}\left(\boldsymbol{x}\right) \right), \quad \boldsymbol{x} \in P^*.
            \end{equation}
            \item \textbf{Convergence indicator ($\mathbf{I_{3}}$)}: The convergence of a candidate solution $x$ is quantified by its the distance to the ideal point $\boldsymbol{z}$. It is calculated by the Eq.~\ref{eq:convergence}.
            \begin{equation}
                \label{eq:convergence}
                d_z\left(\boldsymbol{x}\right)=\left\|\tilde{\hat{\boldsymbol{F}}} \left(\boldsymbol{x}\right)-\boldsymbol{z}\right\|,
                \quad \boldsymbol{x} \in P^*.
            \end{equation}
            where $\boldsymbol{z}=(z_1, z_2, \ldots, z_m)$, $m$ is the number of objectives. Note that $\boldsymbol{z}$ is defined as the minimum objective value across all data points in $\mathcal{D}$ for each objective. The convergence indicator is computed using Eq.~\ref{eq:I3}. A normalization operation is first applied to $d_{z}$, and the resulting value is subsequently negated to obtain the final $\mathbf{I_{3}}$.
            \begin{equation}
                \label{eq:I3}
                \mathbf{I}_3 = -\mathrm{Normalize}\left( d_{z}\left(\boldsymbol{x}\right) \right), \quad \boldsymbol{x} \in P^*.
            \end{equation}
        \end{enumerate}

        \section{EXPERIMENTAL STUDIES}\label{Sec_IV}
        \par To verify the performance of the proposed CI-EMO, comparative studies are conducted by comparing CI-EMO with five state-of-the-art surrogate-assisted MOEAs on multi- and many-objective benchmarks. Compared algorithms include K-RVEA~\citep{Chugh2016-KRVEA}, KTA2~\citep{Song2021-KTA2}, EMMOEA~\citep{Qin2023-EMMOEA}, DirHV-EGO~\citep{Zhao2023-DirHVEGO}, and R2/D-EGO~\citep{Zhao2024-R2DEGO}. Then several experiments were conducted to analyze the effectiveness of CI-EMO, including component analysis of composite indicator, ablation studies, and sampling number analysis. All algorithms are implemented in MATLAB using the PlatEMO platform~\citep{Tian2017-platemo}. To guarantee consistency and reliability, all experiments are conducted on identical hardware specifications (CPU: Intel(R) Core(TM) i9-14900KF, Memory: 128GB).

        \subsection{Experimental Settings}\label{Sec_IV_A}
        \begin{enumerate}
        \item \textbf{Benchmarks Problems}: The DTLZ~\citep{Deb2002-DTLZ}, ZDT~\citep{Deb1999-ZDT}, and MaF~\citep{Cheng2017-benchmark} test suites are selected as the benchmark functions for our research. Referring to~\citep{Zhao2023-DirHVEGO} for EMOPs, the number of decision variables ($d$) is 8 for the two-objective DTLZ and ZDT test suites and 6 for the three-objective DTLZ test suite. Referring to~\citep{Qin2023-EMMOEA} for EMaOPs, algorithms are tested on 3, 5, 10-objective MaF problems, where the number of decision variables is 10 for all test problems except for the 10-objective MaF11 problem whose dimension is set to 11.
        
        \item \textbf{Algorithm Setting}: For all experiments, the number of initial samples and $\mathcal{N}_0$ is set to $11d-1$. $\mathcal{N}_0$ is set to 100 when $11d-1 > 100$. The termination condition is defined by a maximum of 200 function evaluations for $m = 2$ and 300 evaluations for $m \geq 3$. In SA-NSGAIII, $T_{max}$ is set to 20, and the reference vector number and population size are the same as $\mathcal{N}_0$. Each experiment is repeated 21 times to ensure statistical robustness. All parameters for the compared algorithms are configured according to the settings specified in the original papers.
        
        \item \textbf{Performance Evaluation of Algorithms}: We adopted inverted generational distance plus (IGD+)~\citep{Ishibuchi2015-IGD+} and hypervolume (HV)~\citep{Zitzler1999-HV} as performance indicators to evaluate the performance of algorithms. Wilcoxon's rank-sum test~\citep{Wilcoxon1970-Wilcoxon} is conducted at a 5\% significance level to assess whether the proposed CI-EMO exhibits statistically significant differences compared to the algorithms under evaluation. Specifically, the symbols ``$-$'', ``$+$'', and ``$\approx$'' indicate the compared algorithm performs significantly worse, better, or similarly to CI-EMO, respectively.
        \end{enumerate}

        \begin{table*}[ht]\scriptsize
		  \centering
		  \caption{\centering IGD+ STATISTIC RESULTS (MEAN AND STANDARD DEVIATION) FOR CI-EMO AND OTHER FIVE SAEAS ALGORITHMS ON BENCHMARK PROBLEMS}
		  \resizebox{\linewidth}{!}{
            \begin{tabular}{|c|c|c|c|c|c|c|c|c|}
            \hline\hline
            Problem & M     & D     & K-RVEA & KTA2  & EMMOEA & DirHV-EGO & R2/D-EGO & CI-EMO \\
            \hline
            \multirow{0.5}[4]{*}{DTLZ1} & 2     & 8     & 5.2037e+1 (1.27e+1) - & 7.6928e+1 (1.96e+1) - & 3.2222e+1 (1.14e+1) $\approx$ & 9.7419e+1 (2.30e+1) - & 8.6564e+1 (2.43e+1) - & \textbf{2.7890e+1 (1.20e+1)} \\
            \cline{2-9}      & 3     & 6     & 2.3932e+1 (6.92e+0) - & 6.5241e+0 (4.35e+0) + & \textbf{6.1265e+0 (3.35e+0) +} & 1.9380e+1 (7.20e+0) - & 1.6883e+1 (6.29e+0) - & 1.0630e+1 (4.37e+0) \\
            \hline
            \multirow{0.5}[4]{*}{DTLZ2} & 2     & 8     & 2.7490e-2 (1.60e-2) - & 5.7773e-3 (1.11e-3) - & 7.1352e-3 (1.41e-3) - & 5.4591e-3 (1.42e-3) - & 7.0648e-3 (7.92e-4) - & \textbf{3.3969e-3 (5.64e-4)} \\
            \cline{2-9}      & 3     & 6     & 3.4851e-2 (1.95e-3) - & 2.2761e-2 (1.20e-3) - & \textbf{2.1081e-2 (1.36e-3) +} & 2.3294e-2 (1.16e-3) - & 3.1952e-2 (8.69e-4) - & 2.1793e-2 (9.32e-4) \\
            \hline
            \multirow{0.5}[4]{*}{DTLZ3} & 2     & 8     & 1.2233e+2 (4.10e+1) - & 1.4747e+2 (4.29e+1) - & \textbf{7.0313e+1 (3.56e+1) +} & 2.4633e+2 (5.48e+1) - & 2.1015e+2 (5.95e+1) - & 8.4950e+1 (2.80e+1) \\
            \cline{2-9}      & 3     & 6     & 6.2172e+1 (1.57e+1) - & \textbf{1.8699e+1 (1.28e+1) +} & 2.0407e+1 (1.34e+1) + & 4.3296e+1 (1.80e+1) $\approx$ & 3.9740e+1 (1.57e+1) $\approx$ & 4.1563e+1 (1.65e+1) \\
            \hline
            \multirow{0.5}[4]{*}{DTLZ4} & 2     & 8     & 3.0160e-1 (9.82e-2) - & 2.0052e-1 (1.17e-1) - & 1.4705e-1 (9.88e-2) $\approx$ & 2.5043e-1 (8.51e-2) - & 2.5603e-1 (1.13e-1) - & \textbf{1.3820e-1 (1.25e-1)} \\
            \cline{2-9}      & 3     & 6     & 1.3520e-1 (5.00e-2) - & 1.1176e-1 (7.02e-2) $\approx$ & 1.0038e-1 (3.99e-2) $\approx$ & 2.8063e-1 (8.70e-2) - & 2.6896e-1 (5.94e-2) - & \textbf{8.5561e-2 (2.57e-2)} \\
            \hline
            \multirow{0.5}[4]{*}{DTLZ5} & 2     & 8     & 2.2626e-2 (4.74e-3) - & 5.2620e-3 (9.26e-4) - & 7.9606e-3 (1.95e-3) - & 5.2266e-3 (5.67e-4) - & 7.1625e-3 (7.17e-4) - & \textbf{3.3558e-3 (3.75e-4)} \\
            \cline{2-9}      & 3     & 6     & 1.6193e-2 (2.48e-3) - & \textbf{2.1671e-3 (1.28e-4) +} & 4.5064e-3 (7.24e-4) - & 5.4613e-3 (3.11e-4) - & 3.4947e-3 (4.45e-4) - & 2.8684e-3 (1.51e-4) \\
            \hline
            \multirow{0.5}[4]{*}{DTLZ6} & 2     & 8     & 2.8764e+0 (4.45e-1) - & 1.5377e+0 (5.87e-1) - & 1.6060e+0 (5.53e-1) - & \textbf{2.0884e-1 (3.24e-1) +} & 3.5322e-1 (4.44e-1) + & 1.0323e+0 (3.12e-1) \\
            \cline{2-9}      & 3     & 6     & 8.1519e-1 (2.50e-1) - & 2.9276e-1 (1.97e-1) + & 3.5025e-1 (2.17e-1) $\approx$ & 4.9665e-2 (2.19e-2) + & \textbf{2.7975e-2 (3.76e-2) +} & 4.1439e-1 (1.98e-1) \\
            \hline
            \multirow{0.5}[4]{*}{DTLZ7} & 2     & 8     & 1.9052e-2 (4.66e-3) - & 3.9345e-2 (1.07e-1) - & 9.2265e-3 (3.55e-3) - & 6.3478e-2 (9.38e-2) - & 1.9993e-2 (8.78e-3) - & \textbf{1.9703e-3 (7.80e-5)} \\
            \cline{2-9}      & 3     & 6     & 4.2637e-2 (2.62e-3) - & 1.4164e-1 (1.93e-1) - & 1.1671e-1 (1.87e-1) - & 4.3595e-2 (2.72e-2) - & 2.9357e-2 (1.27e-3) - & \textbf{2.2758e-2 (7.66e-4)} \\
            \hline
            ZDT1  & 2     & 8     & 1.6578e-2 (2.19e-3) - & 6.6846e-3 (1.75e-3) - & 3.0837e-2 (2.42e-2) - & 4.2941e-3 (3.48e-4) - & 3.0894e-3 (9.02e-4) $\approx$ & \textbf{2.9650e-3 (2.92e-4)} \\
            \hline
            ZDT2  & 2     & 8     & 2.5248e-2 (2.02e-2) - & 5.2519e-3 (6.48e-4) - & 6.3020e-3 (5.35e-3) - & 2.9172e-3 (1.65e-4) - & 3.6607e-3 (5.31e-4) - & \textbf{2.6868e-3 (1.75e-4)} \\
            \hline
            ZDT3  & 2     & 8     & 1.6831e-2 (5.99e-3) - & 5.8356e-2 (1.10e-1) - & 1.0483e-1 (1.43e-1) - & 3.6427e-2 (6.79e-2) - & 3.6773e-3 (7.16e-4) - & \textbf{2.6893e-3 (4.57e-4)} \\
            \hline
            ZDT4  & 2     & 8     & 2.8346e+1 (1.19e+1) $\approx$ & 2.8620e+1 (8.50e+0) $\approx$ & \textbf{2.1725e+1 (8.52e+0) $\approx$} & 4.8612e+1 (1.11e+1) - & 4.5933e+1 (1.13e+1) - & 2.4561e+1 (9.54e+0) \\
            \hline
            ZDT6  & 2     & 8     & 9.3446e-1 (1.99e-1) - & 4.8135e-1 (1.84e-1) $\approx$ & 4.8224e-1 (2.67e-1) $\approx$ & 1.6881e-1 (1.14e-1) + & \textbf{1.6239e-1 (1.15e-1) +} & 6.0611e-1 (2.52e-1) \\
            \hline
            \multicolumn{3}{|c|}{+/-/$\approx$} & 0/18/1 & 4/12/3 & 4/9/6 & 3/15/1 & 3/14/2 &  \\
            \hline\hline
            \end{tabular}%
            \label{tab:SAEAs}%
            }
	\end{table*}%

        \begin{figure*}[ht]
    		\centering
                    \resizebox{\linewidth}{!}{
    			\includegraphics[scale=0.2]{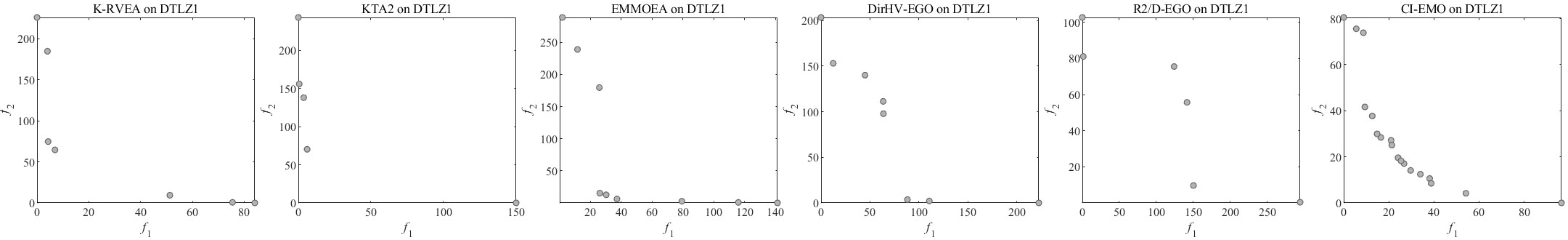}
                    }
                    \resizebox{\linewidth}{!}{
                    \includegraphics[scale=0.2]{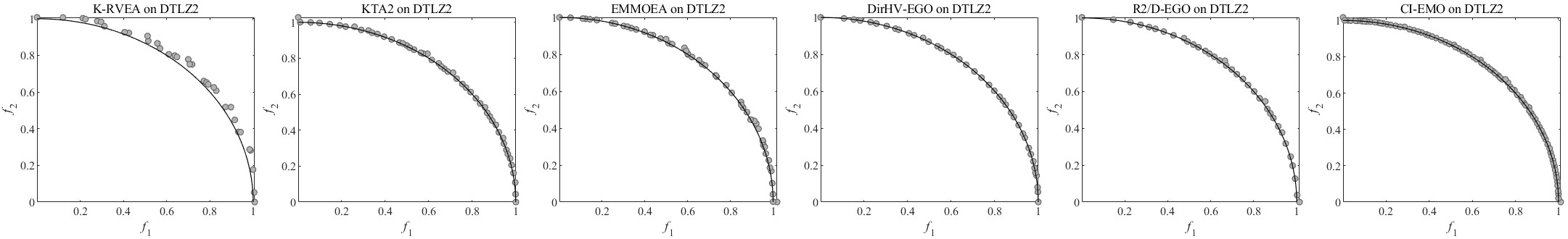}
                    }
                    \resizebox{\linewidth}{!}{
                    \includegraphics[scale=0.2]{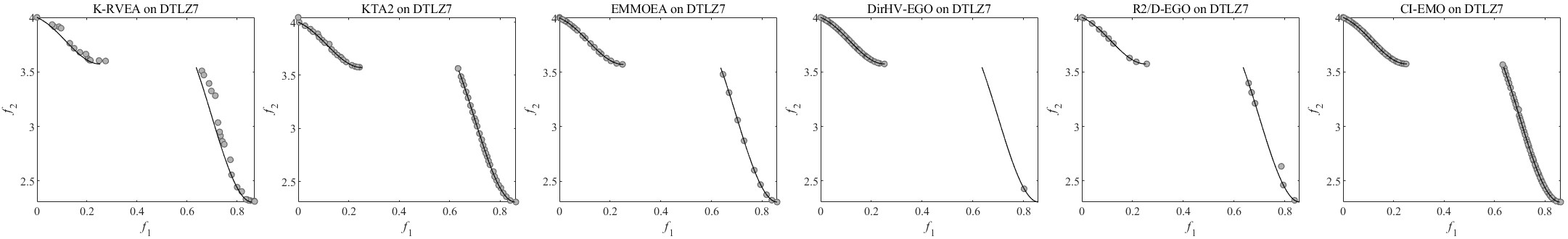}
                    }
                    \resizebox{\linewidth}{!}{
                    \includegraphics[scale=0.24]{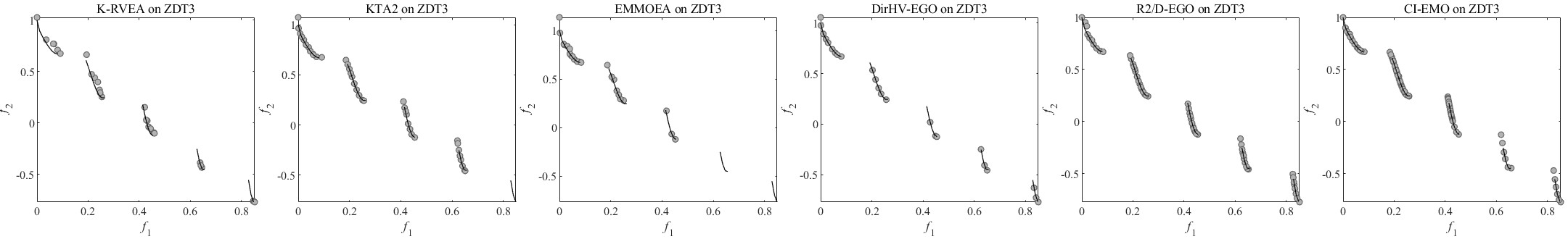}
                    }
    		\caption{The nondominated solutions obtained six algorithms under comparison on 2-objective DTLZ1-2, DTLZ7, and ZDT3.}
    		\label{Figure_4_SAEAs}
	\end{figure*}

        \begin{figure*}[ht]
    		\centering
                    \begin{minipage}{.25\textwidth}
                        \centering
                        \includegraphics[width=\textwidth]{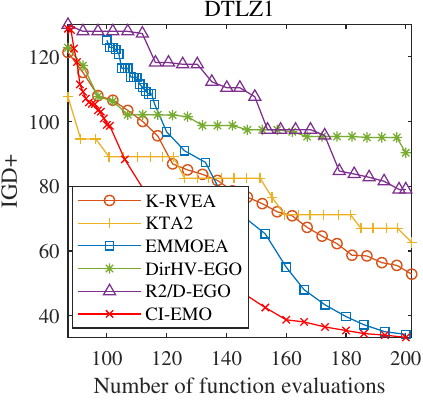}
                    \end{minipage}\hfill
                    \begin{minipage}{.25\textwidth}
                        \centering
                        \includegraphics[width=\textwidth]{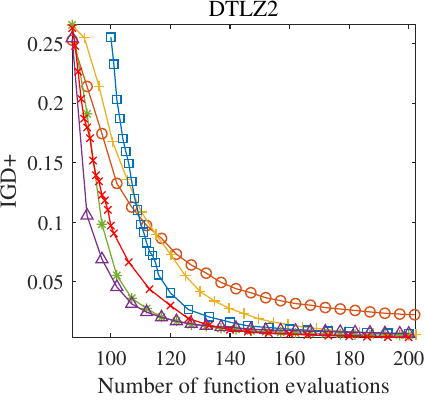}
                    \end{minipage}\hfill
                    \begin{minipage}{.25\textwidth}
                        \centering
                        \includegraphics[width=\textwidth]{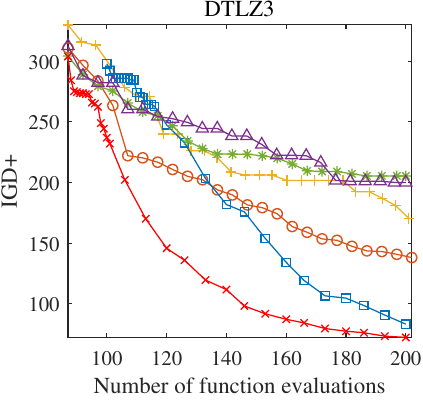}
                    \end{minipage}\hfill
                    \begin{minipage}{.25\textwidth}
                        \centering
                        \includegraphics[width=\textwidth]{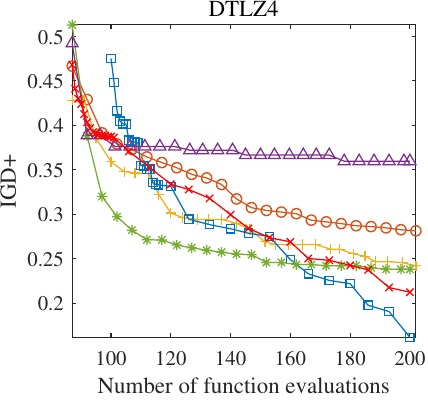}
                    \end{minipage}\hfill
                    \begin{minipage}{.25\textwidth}
                        \centering
                        \includegraphics[width=\textwidth]{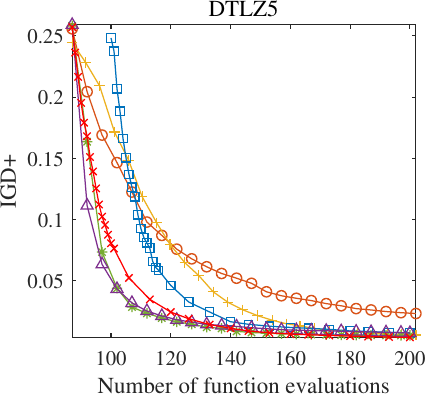}
                    \end{minipage}\hfill
                    \begin{minipage}{.25\textwidth}
                        \centering
                        \includegraphics[width=\textwidth]{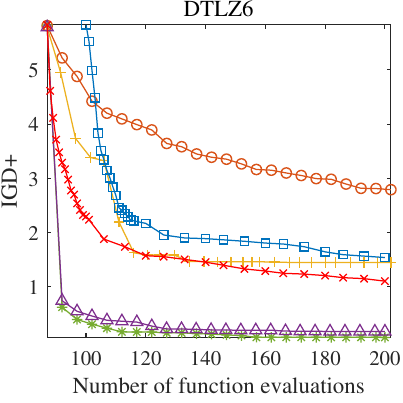}
                    \end{minipage}\hfill
                    \begin{minipage}{.25\textwidth}
                        \centering
                        \includegraphics[width=\textwidth]{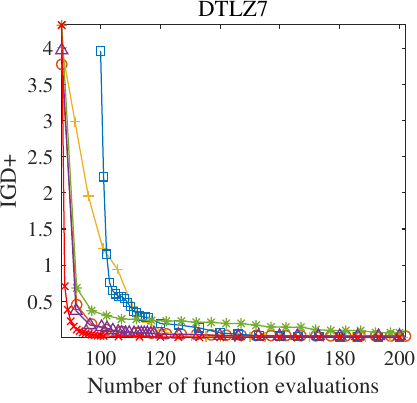}
                    \end{minipage}\hfill
                    \begin{minipage}{.25\textwidth}
                        \centering
                        \includegraphics[width=\textwidth]{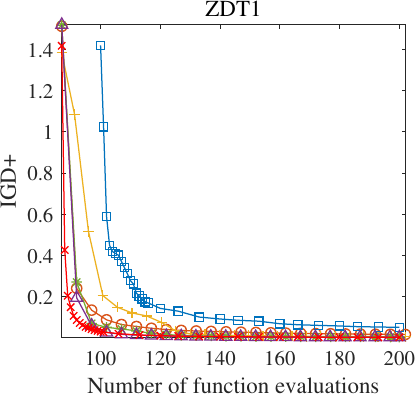}
                    \end{minipage}\hfill
                    \begin{minipage}{.25\textwidth}
                        \centering
                        \includegraphics[width=\textwidth]{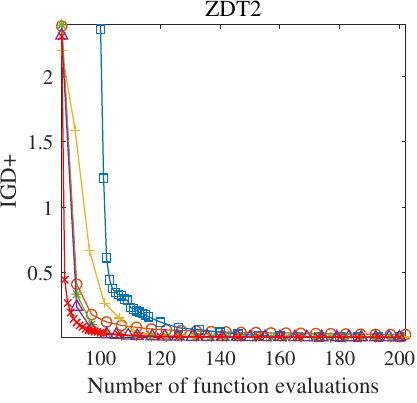}
                    \end{minipage}\hfill
                    \begin{minipage}{.25\textwidth}
                        \centering
                        \includegraphics[width=\textwidth]{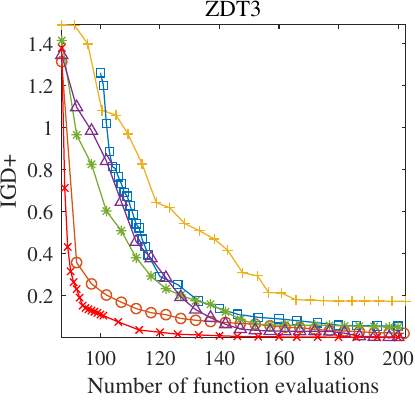}
                    \end{minipage}\hfill
                    \begin{minipage}{.25\textwidth}
                        \centering
                        \includegraphics[width=\textwidth]{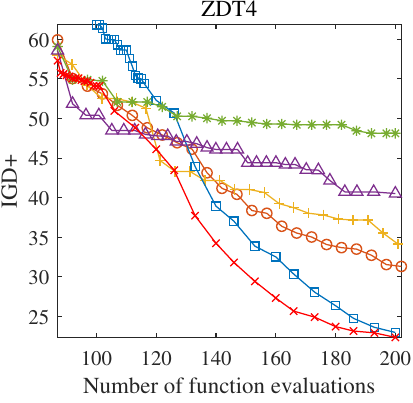}
                    \end{minipage}\hfill
                    \begin{minipage}{.25\textwidth}
                        \centering
                        \includegraphics[width=\textwidth]{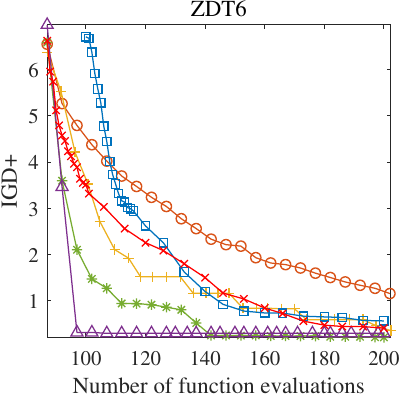}
                    \end{minipage}\hfill
    		\caption{Convergence profiles based on the IGD+ values of six algorithms under comparison on 2-objective DTLZ and ZDT problems.}
    		\label{Figure_6_convergence}
	\end{figure*}
        
        \subsection{Comparison with state-of-the-art SAEAs}\label{Sec_IV_D}
        \par To evaluate the effectiveness of CI-EMO, we compared CI-EMO against five state-of-the-art expensive multi-objective optimization algorithms. The experimental results are summarized in Table~\ref{tab:SAEAs} and~S-I. Table~\ref{tab:SAEAs} reports the mean IGD+ values and standard deviations (std.) of the approximation Pareto-optimal solutions obtained by the six algorithms. From these results, it is clear that CI-EMO achieves the best mean IGD+ values on two-objective problems such as DTLZ1-2, DTLZ4-5, DTLZ7, ZDT1-3, as well as three-objective problems including DTLZ4 and DTLZ7. CI-EMO outperforms the compared algorithms on the majority of test problems, and CI-EMO significantly outperforms K-RVEA on 18 out of 19 problems expected ZDT4. CI-EMO performs better than K-RVEA, KTA2, EMMOEA, DirHV-EGO, and R2/D-EGO on 18, 12, 9, 15, and 14 out of 19 problems, respectively, while being inferior to these algorithms on only 0, 4, 4, 3, and 3 problems, respectively. Table~S-I shows the mean HV values and standard deviations of the approximation Pareto-optimal solutions obtained by the six algorithms. CI-EMO performs better than K-RVEA, KTA2, EMMOEA, DirHV-EGO, and R2/D-EGO on 13, 7, 7, 10, and 10 out of 19 problems, while it is inferior to them only on 0, 1, 0, 3, and 3 problems, respectively. Its consistent superiority in terms of both IGD+ and HV metrics across a range of benchmark problems demonstrates its effectiveness and robustness. Thus, we can conclude that CI-EMO demonstrates strong competitiveness in solving expensive multi-objective optimization problems.
        \par To gain further insights, we analyzed the performance of CI-EMO on problems with varying Pareto front characteristics. CI-EMO performs effectively on convex Pareto fronts (e.g., ZDT1, DTLZ2), concave Pareto fronts (e.g., ZDT2), and discontinuous Pareto fronts (e.g., ZDT3, DTLZ7). As illustrated in Fig.~\ref{Figure_4_SAEAs}, CI-EMO demonstrates superior performance in identifying well-distributed Pareto solution sets compared to other advanced methods, particularly for problems with discontinuous Pareto fronts, such as DTLZ7 and ZDT3. However, we also observe that complex problems with multimodal functional landscapes, such as ZDT4 and DTLZ3, pose significant challenges in finding the Pareto Front within a limited number of fitness evaluation budgets. IGD+ values are over 20 on ZDT4. The non-dominated solution set is still far from the Pareto frontier. The advantage of CI-EMO sampling that takes distribution into account is not exploited, resulting in CI-EMO achieving second place on this problem rather than the best result. To demonstrate the performance of CI-EMO more convincingly, we plot the convergence curves of the IGD+ values of the six algorithms on the 2-objective DTLZ, ZDT test set in Fig.~\ref{Figure_6_convergence}, which are averaged over 21 independent runs. The legend is shown in the DTZL1 subfigure. From Fig.~\ref{Figure_6_convergence}, we can see that when the number of function evaluations is less than 160, the proposed CI-EMO method converges relatively much faster for many problems, such as DTLZ1, DTLZ3, DTLZ5, DTLZ7, and ZDT1-4.

        \begin{table*}[h]\scriptsize
		  \centering
		  \caption{\centering IGD+ STATISTIC RESULTS (MEAN AND STANDARD DEVIATION) FOR CI-EMO AND OTHER FIVE SAEAS ALGORITHMS ON MaF PROBLEMS}
            \resizebox{\linewidth}{!}{
            \begin{tabular}{|c|c|c|c|c|c|c|c|c|}
            \hline\hline
            Problem & M     & D     & KRVEA & KTA2  & EMMOEA & DirHVEGO & R2DEGO & CI-EMO \\
            \hline
            \multirow{3}[2]{*}{MaF1} & 3     & 10    & 5.6870e-2 (3.81e-3) - & 3.5729e-2 (5.08e-3) - & 3.2367e-2 (1.84e-3) - & 5.0724e-2 (2.42e-3) - & 5.1976e-2 (2.90e-3) - & \textbf{2.9443e-2 (1.36e-3)} \\
                  & 5     & 10    & 1.1066e-1 (1.07e-2) - & 1.1469e-1 (1.02e-2) - & 1.0234e-1 (1.02e-2) - & 1.2893e-1 (2.89e-3) - & 1.0957e-1 (2.31e-3) - & \textbf{9.2007e-2 (2.83e-3)} \\
                  & 10    & 10    & 2.9938e-1 (1.15e-2) - & 2.2126e-1 (1.56e-2) - & 2.0215e-1 (8.20e-3) - & 2.8917e-1 (1.39e-2) - & 1.9254e-1 (1.72e-2) - & \textbf{1.6215e-1 (2.39e-3)} \\
            \hline
            \multirow{3}[2]{*}{MaF2} & 3     & 10    & 2.8508e-2 (3.76e-3) - & 2.2128e-2 (1.42e-3) - & 2.2778e-2 (2.56e-3) - & 1.9072e-2 (4.19e-4) - & 2.0915e-2 (2.96e-4) - & \textbf{1.7840e-2 (1.06e-3)} \\
                  & 5     & 10    & 5.9264e-2 (1.55e-3) - & 5.8337e-2 (1.14e-3) - & 5.5798e-2 (1.68e-3) - & 5.9942e-2 (9.37e-4) - & 5.9850e-2 (1.44e-3) - & \textbf{5.3672e-2 (1.88e-3)} \\
                  & 10    & 10    & 1.6731e-1 (1.15e-2) - & 1.1191e-1 (3.27e-3) - & 1.6755e-1 (6.20e-3) - & 1.5873e-1 (4.02e-3) - & 1.7162e-1 (8.69e-3) - & \textbf{1.0356e-1 (3.12e-3)} \\
            \hline
            \multirow{3}[2]{*}{MaF3} & 3     & 10    & 1.2793e+5 (5.75e+4) - & 3.5680e+5 (5.09e+5) $\approx$ & 4.8569e+5 (3.11e+5) - & 4.5166e+5 (1.90e+5) - & 4.3561e+5 (2.29e+5) - & \textbf{7.9113e+4 (2.69e+4)} \\
                  & 5     & 10    & \textbf{5.7083e+4 (3.42e+4) $\approx$} & 1.7770e+5 (1.72e+5) $\approx$ & 2.0127e+5 (1.11e+5) - & 1.7327e+5 (1.39e+5) - & 1.9853e+5 (8.90e+4) - & 8.3258e+4 (6.85e+4) \\
                  & 10    & 10    & 4.2330e+0 (5.92e+0) $\approx$ & \textbf{7.6104e-1 (1.01e+0) +} & 3.0255e+0 (4.17e+0) $\approx$ & 2.4092e+0 (3.45e+0) + & 1.0624e+0 (1.07e+0) + & 4.4752e+0 (5.00e+0) \\
            \hline
            \multirow{3}[2]{*}{MaF4} & 3     & 10    & 8.3810e+2 (2.61e+2) - & 3.9096e+2 (1.93e+2) $\approx$ & \textbf{2.4332e+2 (9.64e+1) +} & 1.1036e+3 (2.33e+2) - & 8.7200e+2 (1.93e+2) - & 3.9818e+2 (9.24e+1) \\
                  & 5     & 10    & 1.3349e+3 (4.24e+2) - & 7.3537e+2 (3.81e+2) $\approx$ & \textbf{5.1640e+2 (2.48e+2) $\approx$} & 1.8811e+3 (9.13e+2) - & 2.2037e+3 (7.11e+2) - & 6.4552e+2 (2.35e+2) \\
                  & 10    & 10    & 8.9324e+1 (1.19e+2) - & 2.7565e+1 (6.80e+0) - & 2.3608e+1 (8.28e+0) $\approx$ & 3.9661e+1 (1.34e+1) - & 3.8717e+1 (8.79e+0) - & \textbf{2.0031e+1 (3.92e+0)} \\
            \hline
            \multirow{3}[2]{*}{MaF5} & 3     & 10    & 1.2048e+0 (3.27e-1) - & 7.7051e-1 (3.77e-1) $\approx$ & \textbf{6.0454e-1 (3.14e-1) +} & 1.8046e+0 (4.79e-1) - & 1.5727e+0 (4.40e-1) - & 8.5404e-1 (3.25e-1) \\
                  & 5     & 10    & 1.9636e+0 (6.07e-1) - & 1.9463e+0 (1.25e+0) $\approx$ & \textbf{1.2938e+0 (2.71e-1) +} & 2.8146e+0 (1.10e+0) - & 1.9879e+0 (4.30e-1) - & 1.4906e+0 (2.57e-1) \\
                  & 10    & 10    & 2.4275e+0 (2.41e+0) - & 2.3355e+0 (1.74e+0) $\approx$ & \textbf{1.4380e+0 (1.40e-1) $\approx$} & 1.6705e+0 (3.28e-1) - & 1.4578e+0 (5.54e-2) $\approx$ & 1.4643e+0 (9.72e-2) \\
            \hline
            \multirow{3}[2]{*}{MaF6} & 3     & 10    & 6.8353e-1 (2.69e-1) - & 1.3683e-1 (2.96e-1) - & \textbf{2.9487e-2 (8.61e-3) +} & 4.9163e-1 (1.80e-1) - & 2.4801e-1 (6.10e-2) - & 5.5014e-2 (2.48e-2) \\
                  & 5     & 10    & 3.5803e-1 (1.55e-1) - & \textbf{3.4018e-2 (1.10e-2) +} & 3.5580e-2 (1.36e-2) + & 1.4198e-1 (4.85e-2) $\approx$ & 1.6012e-1 (4.52e-2) $\approx$ & 1.7337e-1 (8.57e-2) \\
                  & 10    & 10    & 1.2010e-2 (7.05e-3) - & 6.8670e-3 (5.18e-3) - & 1.1639e-2 (2.75e-3) - & 2.7244e-2 (6.00e-3) - & 8.8228e-3 (1.16e-3) - & \textbf{3.6029e-3 (5.06e-4)} \\
            \hline
            \multirow{3}[2]{*}{MaF7} & 3     & 10    & 6.9067e-2 (5.46e-3) - & 1.6480e-1 (1.89e-1) - & 1.3996e-1 (2.21e-1) - & 9.3166e-2 (3.84e-2) - & 4.0208e-2 (3.63e-3) - & \textbf{2.5675e-2 (1.53e-3)} \\
                  & 5     & 10    & 3.4270e-1 (4.16e-2) - & 3.5105e-1 (1.64e-1) - & 6.5229e-1 (2.80e-1) - & 6.0425e-1 (1.85e-1) - & \textbf{1.6954e-1 (1.92e-2) +} & 2.0703e-1 (1.43e-2) \\
                  & 10    & 10    & \textbf{9.5581e-1 (4.44e-2) +} & 1.0786e+0 (1.82e-1) + & 1.9665e+0 (4.03e-1) - & 2.0222e+0 (5.79e-1) - & 9.9870e-1 (4.50e-2) + & 1.4772e+0 (1.96e-1) \\
            \hline
            \multirow{3}[2]{*}{MaF10} & 3     & 10    & 1.7662e+0 (9.48e-2) $\approx$ & \textbf{1.6581e+0 (1.76e-1) $\approx$} & 1.6717e+0 (9.32e-2) + & 1.9506e+0 (8.74e-2) - & 1.9563e+0 (7.19e-2) - & 1.7185e+0 (1.01e-1) \\
                  & 5     & 10    & 2.2164e+0 (5.94e-2) - & 2.0923e+0 (1.97e-1) $\approx$ & 2.0716e+0 (1.92e-1) $\approx$ & 2.3081e+0 (7.63e-2) - & 2.3438e+0 (5.65e-2) - & \textbf{2.0469e+0 (1.13e-1)} \\
                  & 10    & 10    & 2.8488e+0 (7.99e-2) - & 1.3457e+0 (8.42e-1) - & 1.2599e+0 (7.00e-1) - & 2.8146e+0 (3.83e-1) - & 2.9298e+0 (7.13e-2) - & \textbf{7.8454e-1 (1.96e-1)} \\
            \hline
            \multirow{3}[2]{*}{MaF11} & 3     & 10    & 3.3240e-1 (1.10e-1) - & 2.3424e-1 (6.74e-2) - & 1.9871e-1 (2.05e-2) - & 2.8044e-1 (3.49e-2) - & 2.4255e-1 (1.82e-2) - & \textbf{1.5600e-1 (2.03e-2)} \\
                  & 5     & 10    & 3.8467e-1 (1.35e-1) - & 3.9002e-1 (7.91e-2) - & 4.1626e-1 (3.36e-1) $\approx$ & 4.2232e-1 (7.57e-2) - & 3.6136e-1 (5.84e-2) - & \textbf{2.7432e-1 (4.61e-2)} \\
                  & 10    & 11    & 5.6374e-1 (3.11e-1) - & 7.8263e-1 (1.13e-1) - & 1.2212e+0 (8.81e-1) - & 4.4496e-1 (9.63e-2) - & 1.0428e+0 (8.27e-1) - & \textbf{3.6306e-1 (5.92e-2)} \\
            \hline
            \multirow{3}[2]{*}{MaF12} & 3     & 10    & 6.2778e-1 (7.76e-2) - & 5.2826e-1 (8.53e-2) $\approx$ & \textbf{4.0613e-1 (6.93e-2) +} & 5.0199e-1 (7.13e-2) $\approx$ & 4.9225e-1 (6.88e-2) $\approx$ & 5.0258e-1 (7.47e-2) \\
                  & 5     & 10    & 1.4551e+0 (1.87e-1) - & 1.1430e+0 (2.06e-1) $\approx$ & \textbf{9.0391e-1 (1.40e-1) +} & 1.2723e+0 (2.99e-1) $\approx$ & 1.2912e+0 (3.64e-1) $\approx$ & 1.1081e+0 (2.59e-1) \\
                  & 10    & 10    & 5.2619e+0 (7.56e-1) - & 3.8566e+0 (4.51e-1) $\approx$ & \textbf{3.5319e+0 (4.50e-1) $\approx$} & 6.2536e+0 (9.76e-1) - & 5.6526e+0 (9.39e-1) - & 3.8377e+0 (6.10e-1) \\
            \hline
            \multirow{3}[2]{*}{MaF13} & 3     & 10    & 2.8980e-1 (4.90e-2) - & 1.4864e-1 (6.06e-2) $\approx$ & 1.1347e-1 (4.11e-2) $\approx$ & 7.4109e-2 (5.38e-3) + & \textbf{4.2852e-2 (2.17e-3) +} & 1.1340e-1 (3.21e-2) \\
                  & 5     & 10    & 2.9703e+1 (2.22e+1) - & \textbf{2.4959e-1 (5.32e-2) +} & 3.7666e-1 (8.06e-2) + & 2.7080e-1 (4.10e-2) + & 2.9853e+0 (3.88e+0) - & 5.4558e-1 (1.20e-1) \\
                  & 10    & 10    & 6.3797e+1 (5.79e+1) - & 4.2424e-1 (1.33e-1) + & 4.8003e-1 (1.46e-1) + & \textbf{2.7382e-1 (4.89e-2) +} & 1.4196e+1 (1.55e+1) - & 7.4485e-1 (3.26e-1) \\
            \hline
            \multicolumn{3}{|c|}{+/-/$\approx$} & 1/29/3 & 5/15/13 & 10/15/8 & 4/26/3 & 4/25/4 &  \\
            \hline\hline
            \end{tabular}%
            }
            \label{tab:SAEAs-MaF}%
        \end{table*}%

        \subsection{Comparison with state-of-the-art SAEAs on Many-objective Problems}\label{Sec_IV_D}
        To further research CI-EMO performance on expensive many-objective problems, we compare algorithms on 3-, 5-, and 10-objective MaF test problems. IGD+ Results are shown in Table~\ref{tab:SAEAs-MaF}. CI-EMO performs better than K-RVEA, KTA2, EMMOEA, DirHV-EGO, and R2/D-EGO on 29, 15, 15, 26, and 25 out of 33 problems, respectively, while being inferior to these algorithms on only 1, 5, 10, 4, and 4 problems, respectively. We find that CI-EMO is most efficient in solving MaF1-3 and MaF7-11. CI-EMO is weaker than EMMOEA on MaF4, MaF6, and MaF12-13 since EMMOEA has an advantage in solving concave PF problems, which has been described on~\citep {Qin2023-EMMOEA}. Table~S-II shows the mean HV values and standard deviations obtained by the six algorithms. CI-EMO performs better than K-RVEA, KTA2, EMMOEA, DirHV-EGO, and R2/D-EGO on 27, 16, 11, 18, and 22 out of 33 problems, while it is inferior to them only on 0, 4, 9, 6, and 3 problems, respectively. Overall, CI-EMO has the best performance, which verifies the effectiveness of composite indicator-based sampling in high-dimensional objective space. Thus, experiments demonstrate CI-EMO not only has strong competitiveness in solving expensive multi-objective optimization problems but also works well on many-objective optimization problems.

        \subsection{Effectiveness of Composite Indicator}\label{Sec_IV_B}
        \par To validate the effectiveness of the composite indicator, this paper designs two types of experiments: one to analyze the properties of each component in the composite indicator, and another to assess the impact of different components of the indicator by ablation experiments. First, CI-EMO is compared with four variants, where $rand$-EMO refers to the method of randomly selecting an individual from candidate population for real fitness evaluation, while the other three variants correspond to using one of the three performance indicators for candidate selection, respectively. ``M'' represents the number of objectives and "D" means the dimension of the problem. $\mathbf{I}_1$-EMO means only $\mathbf{I}_1$ is used in the composite indicator. The experimental results are presented in Table~\ref{tab:CI-EMO-S}. As shown in the table, CI-EMO significantly outperforms $rand$-EMO on 13 test problems. Furthermore, the algorithm's performance is inferior to CI-EMO when only a single indicator ($\mathbf{I}_1$, $\mathbf{I}_2$, or $\mathbf{I}_3$) is used. It is evident that using different indicators produces significantly varied results across different test problems, as shown in Fig.~S-1. Compared with single sampling indicator-based variants, CI-EMO has better robustness. Among the three variants that use a single performance indicator, $\mathbf{I}_1$-EMO performs the best. It shows no significant difference compared to CI-EMO in 14 test cases and is significantly worse in five test problems. $\mathbf{I}_2$-EMO, performs significantly worse than CI-EMO in 9 test cases. It performs well on the ZDT test set but underperforms on the DTLZ test set. $\mathbf{I}_3$-EMO exhibits the poorest performance; except for DTLZ1 and DTLZ3, it is significantly worse than CI-EMO in most of the problems. As shown in Fig.~S-1, if only $\mathbf{I}_3$ is used, solutions on PF will lack diversity, selection of solutions based on the convergence indicator is biased towards certain regions of the PF as results on DTLZ2 and ZDT3. In addition, the algorithm benefits from multiple iterations of surrogate model-assisted NSGA-III during the generation process of the candidate population. Individuals in the candidate population have exhibited convergence. Therefore, $\mathbf{I}_1$-EMO and $\mathbf{I}_2$-EMO have better performance than $\mathbf{I}_3$-EMO.

        \par On the other hand, ablation experiments have been conducted. We delete one metric from the composite indicator in each variant. Results have been shown in Table~\ref{tab:CI-EMO-SS}. CI-EMO-no-$\mathbf{I}_1$ means that $\mathbf{I}_2$ and $\mathbf{I}_3$ are used in the composite indicator, and $\mathbf{I}_1$ is deleted from the composite indicator. We can see that no matter which of the three metrics is deleted, the performance of the algorithm will drop significantly. Among them, deleting $\mathbf{I}_3$ has the least impact on the proposed algorithm, and the effect of CI-EMO-no-$\mathbf{I}_3$ is better on some problems, such as ZDT1-3.
        
        \begin{table*}[h]\scriptsize
		  \centering
		  \caption{\centering IGD+ STATISTIC RESULTS (MEAN AND STANDARD DEVIATION) FOR CI-EMO AND OTHER FOUR VARIANTS ON BENCHMARK PROBLEMS}
		\resizebox{0.85\linewidth}{!}{
            \begin{tabular}{|c|c|c|c|c|c|c|c|}
            \hline\hline
            Problem & M     & D     & $rand$-EMO & $\mathbf{I}_1$-EMO & $\mathbf{I}_2$-EMO & $\mathbf{I}_3$-EMO & CI-EMO \\
            \hline
            \multirow{0.5}[4]{*}{DTLZ1} & 2     & 8     & 5.2750e+1 (1.76e+1) - & \textbf{2.9005e+1 (1.02e+1) $\approx$} & 7.5850e+1 (1.09e+1) - & 3.2996e+1 (2.51e+1) $\approx$ & 3.0662e+1 (8.78e+0) \\
            \cline{2-8}      & 3     & 6     & 9.8729e+0 (4.29e+0) $\approx$ & 1.2067e+1 (4.62e+0) $\approx$ & 2.3649e+1 (5.58e+0) - & \textbf{8.2158e+0 (4.11e+0) $\approx$} & 9.4145e+0 (4.68e+0) \\
            \hline
            \multirow{0.5}[4]{*}{DTLZ2} & 2     & 8     & 7.4527e-3 (2.27e-3) - & 3.8079e-3 (1.68e-3) $\approx$ & 4.1698e-3 (1.42e-3) - & 1.1791e-1 (5.86e-2) - & \textbf{3.4063e-3 (4.00e-4)} \\
            \cline{2-8}      & 3     & 6     & 3.1120e-2 (1.76e-3) - & 2.3325e-2 (1.55e-3) - & 2.2279e-2 (1.07e-3) $\approx$ & 1.6252e-1 (3.50e-2) - & \textbf{2.2150e-2 (1.83e-3)} \\
            \hline
            \multirow{0.5}[4]{*}{DTLZ3} & 2     & 8     & 1.2074e+2 (4.26e+1) - & 7.0933e+1 (2.29e+1) $\approx$ & 1.4508e+2 (1.85e+1) - & \textbf{5.6607e+1 (2.17e+1) +} & 7.6125e+1 (2.95e+1) \\
            \cline{2-8}      & 3     & 6     & 3.1313e+1 (1.42e+1) $\approx$ & 3.4914e+1 (1.62e+1) $\approx$ & 6.0286e+1 (9.73e+0) - & \textbf{2.6321e+1 (1.43e+1) +} & 4.0630e+1 (1.72e+1) \\
            \hline
            \multirow{0.5}[4]{*}{DTLZ4} & 2     & 8     & 2.1532e-1 (1.19e-1) $\approx$ & 2.1674e-1 (1.28e-1) $\approx$ & \textbf{1.8012e-1 (1.36e-1) $\approx$} & 3.1581e-1 (9.10e-2) - & 1.9344e-1 (1.35e-1) \\
            \cline{2-8}      & 3     & 6     & 1.1794e-1 (6.39e-2) $\approx$ & 1.1879e-1 (5.52e-2) $\approx$ & 1.2503e-1 (4.67e-2) - & 3.3192e-1 (1.32e-1) - & \textbf{9.2780e-2 (3.91e-2)} \\
            \hline
            \multirow{0.5}[4]{*}{DTLZ5} & 2     & 8     & 6.7024e-3 (2.65e-3) - & 3.7326e-3 (5.39e-4) $\approx$ & 4.1974e-3 (1.18e-3) - & 1.1086e-1 (4.80e-2) - & \textbf{3.4790e-3 (4.10e-4)} \\
            \cline{2-8}      & 3     & 6     & 7.6767e-3 (1.21e-3) - & 3.5436e-3 (5.04e-4) - & 3.1041e-3 (1.89e-4) - & 9.0269e-2 (2.06e-2) - & \textbf{2.9033e-3 (1.68e-4)} \\
            \hline
            \multirow{0.5}[4]{*}{DTLZ6} & 2     & 8     & 1.2537e+0 (3.87e-1) - & 1.0621e+0 (2.94e-1) $\approx$ & 1.1651e+0 (3.32e-1) - & 1.2111e+0 (4.35e-1) - & \textbf{8.8361e-1 (3.18e-1)} \\
            \cline{2-8}      & 3     & 6     & \textbf{3.1360e-1 (1.89e-1) $\approx$} & 3.3204e-1 (1.34e-1) $\approx$ & 3.6064e-1 (1.81e-1) $\approx$ & 4.5097e-1 (2.60e-1) $\approx$ & 3.2192e-1 (1.67e-1) \\
            \hline
            \multirow{0.5}[4]{*}{DTLZ7} & 2     & 8     & 4.9261e-3 (8.55e-4) - & 2.3362e-3 (1.21e-4) - & 1.9942e-3 (9.07e-5) $\approx$ & 6.8760e-1 (4.74e-2) - & \textbf{1.9779e-3 (6.93e-5)} \\
            \cline{2-8}      & 3     & 6     & 4.1313e-2 (1.42e-3) - & 2.5958e-2 (8.88e-4) - & 2.2888e-2 (4.71e-4) $\approx$ & 1.1527e+0 (3.70e-1) - & \textbf{2.2759e-2 (6.08e-4)} \\
            \hline
            ZDT1  & 2     & 8     & 6.1809e-3 (9.91e-4) - & 2.8999e-3 (1.19e-4) $\approx$ & \textbf{2.8739e-3 (2.75e-4) $\approx$} & 4.9748e-1 (1.28e-1) - & 2.9639e-3 (3.48e-4) \\
            \hline
            ZDT2  & 2     & 8     & 6.2232e-3 (2.13e-3) - & 2.7064e-3 (1.54e-4) $\approx$ & \textbf{2.6581e-3 (1.63e-4) $\approx$} & 8.7773e-3 (1.21e-2) - & 2.6829e-3 (1.33e-4) \\
            \hline
            ZDT3  & 2     & 8     & 7.5666e-3 (7.32e-3) - & 6.5624e-2 (2.68e-2) - & \textbf{1.8876e-3 (2.20e-4) +} & 4.4707e-1 (8.18e-2) - & 2.6870e-3 (5.41e-4) \\
            \hline
            ZDT4  & 2     & 8     & 2.3933e+1 (1.16e+1) $\approx$ & \textbf{2.0954e+1 (9.67e+0) $\approx$} & 2.3737e+1 (9.33e+0) $\approx$ & 3.1791e+1 (1.12e+1) - & 2.2829e+1 (1.14e+1) \\
            \hline
            ZDT6  & 2     & 8     & 1.4787e+0 (1.10e+0) - & 4.1541e-1 (1.42e-1) $\approx$ & \textbf{3.9620e-1 (1.47e-1) $\approx$} & 4.4837e+0 (9.67e-1) - & 4.1481e-1 (1.10e-1) \\
            \hline
            \multicolumn{3}{|c|}{+/-/$\approx$} & 0/13/6 & 0/5/14 & 1/9/9 & 2/14/3 &  \\
            \hline\hline
            \end{tabular}%
            \label{tab:CI-EMO-S}%
        }
	\end{table*}%

        \begin{table*}[h]\scriptsize
		  \centering
		  \caption{\centering IGD+ STATISTIC RESULTS (MEAN AND STANDARD DEVIATION) FOR CI-EMO AND OTHER THREE VARIANTS ON BENCHMARK PROBLEMS}
		\resizebox{0.72\linewidth}{!}{
            \begin{tabular}{|c|c|c|c|c|c|c|}
            \hline\hline
            Problem & M     & D     & CI-EMO-no-$\mathbf{I}_1$ & CI-EMO-no-$\mathbf{I}_2$ & CI-EMO-no-$\mathbf{I}_3$ & CI-EMO \\
            \hline
            \multirow{0.5}[4]{*}{DTLZ1} & 2     & 8     & 4.3181e+1 (1.95e+1) $\approx$ & \textbf{2.3327e+1 (8.21e+0) +} & 3.8788e+1 (1.19e+1) - & 3.0662e+1 (8.78e+0) \\
            \cline{2-7}      & 3     & 6     & 1.0785e+1 (5.22e+0) $\approx$ & 9.4244e+0 (4.03e+0) $\approx$ & 1.6333e+1 (6.29e+0) - & \textbf{9.4145e+0 (4.68e+0)} \\
            \hline
            \multirow{0.5}[4]{*}{DTLZ2} & 2     & 8     & 3.5893e-3 (4.76e-4) $\approx$ & 4.6717e-3 (2.27e-3) - & 3.4209e-3 (2.85e-4) $\approx$ & \textbf{3.4063e-3 (4.00e-4)} \\
            \cline{2-7}      & 3     & 6     & 2.2241e-2 (1.08e-3) $\approx$ & 2.2206e-2 (1.41e-3) $\approx$ & \textbf{2.2009e-2 (9.31e-4) $\approx$} & 2.2150e-2 (1.83e-3) \\
            \hline
            \multirow{0.5}[4]{*}{DTLZ3} & 2     & 8     & 9.4263e+1 (2.51e+1) - & \textbf{6.9811e+1 (2.77e+1) $\approx$} & 1.0721e+2 (2.93e+1) - & 7.6125e+1 (2.95e+1) \\
            \cline{2-7}      & 3     & 6     & \textbf{2.9583e+1 (1.45e+1) +} & 3.1562e+1 (2.05e+1) $\approx$ & 5.4427e+1 (1.35e+1) - & 4.0630e+1 (1.72e+1) \\
            \hline
            \multirow{0.5}[4]{*}{DTLZ4} & 2     & 8     & 2.6038e-1 (1.22e-1) $\approx$ & 1.9603e-1 (1.46e-1) $\approx$ & 2.0650e-1 (1.20e-1) $\approx$ & \textbf{1.9344e-1 (1.35e-1)} \\
            \cline{2-7}      & 3     & 6     & 1.1061e-1 (4.20e-2) $\approx$ & 1.0020e-1 (4.58e-2) $\approx$ & 1.2195e-1 (5.97e-2) $\approx$ & \textbf{9.2780e-2 (3.91e-2)} \\
            \hline
            \multirow{0.5}[4]{*}{DTLZ5} & 2     & 8     & 3.6299e-3 (3.57e-4) $\approx$ & 4.1419e-3 (1.17e-3) - & \textbf{3.4655e-3 (5.08e-4) $\approx$} & 3.4790e-3 (4.10e-4) \\
            \cline{2-7}      & 3     & 6     & 3.0425e-3 (2.66e-4) - & 2.9644e-3 (2.24e-4) $\approx$ & 3.1520e-3 (1.97e-4) - & \textbf{2.9033e-3 (1.68e-4)} \\
            \hline
            \multirow{0.5}[4]{*}{DTLZ6} & 2     & 8     & 1.0452e+0 (3.03e-1) $\approx$ & 1.1750e+0 (3.93e-1) - & 1.2064e+0 (2.99e-1) - & \textbf{8.8361e-1 (3.18e-1)} \\
            \cline{2-7}      & 3     & 6     & 3.8907e-1 (1.87e-1) $\approx$ & 4.1342e-1 (1.16e-1) - & 3.7544e-1 (1.49e-1) $\approx$ & \textbf{3.2192e-1 (1.67e-1)} \\
            \hline
            \multirow{0.5}[4]{*}{DTLZ7} & 2     & 8     & 2.2365e-3 (1.55e-4) - & 2.2058e-3 (1.11e-4) - & 2.0711e-3 (1.19e-4) - & \textbf{1.9779e-3 (6.93e-5)} \\
            \cline{2-7}      & 3     & 6     & 2.5277e-2 (9.88e-4) - & 2.5093e-2 (1.08e-3) - & \textbf{2.2716e-2 (7.11e-4) $\approx$} & 2.2759e-2 (6.08e-4) \\
            \hline
            ZDT1  & 2     & 8     & 3.6589e-3 (1.25e-3) $\approx$ & 3.4208e-3 (2.79e-4) - & \textbf{2.7481e-3 (2.35e-4) +} & 2.9639e-3 (3.48e-4) \\
            \hline
            ZDT2  & 2     & 8     & 4.5818e-3 (7.00e-3) - & 3.0322e-3 (1.90e-4) - & \textbf{2.5798e-3 (1.37e-4) +} & 2.6829e-3 (1.33e-4) \\
            \hline
            ZDT3  & 2     & 8     & \textbf{2.0387e-3 (3.20e-4) +} & 9.1221e-2 (3.25e-2) - & 2.2150e-3 (2.85e-4) + & 2.6870e-3 (5.41e-4) \\
            \hline
            ZDT4  & 2     & 8     & 2.5495e+1 (1.32e+1) $\approx$ & \textbf{2.1654e+1 (1.04e+1) $\approx$} & 2.1911e+1 (9.66e+0) $\approx$ & 2.2829e+1 (1.14e+1) \\
            \hline
            ZDT6  & 2     & 8     & 5.9842e-1 (2.26e-1) - & 6.1770e-1 (3.30e-1) - & 4.3090e-1 (2.21e-1) $\approx$ & \textbf{4.1481e-1 (1.10e-1)} \\
            \hline
            \multicolumn{3}{|c|}{+/-/$\approx$} & 2/6/11 & 1/10/8 & 3/7/9 &  \\
            \hline\hline
            \end{tabular}%
            \label{tab:CI-EMO-SS}%
        }
	\end{table*}%

        \begin{table}[h]\scriptsize
		  \centering
		  \caption{\centering IGD+ STATISTIC RESULTS (MEAN AND STANDARD DEVIATION) FOR CI-EMO AND THE VARIANT WITHOUT NORMALIZATION}
            \resizebox{0.7\linewidth}{!}{
            \begin{tabular}{|c|c|c|c|c|}
            \hline\hline
            Problem & M     & D     & CI-EMO-no-Norm & CI-EMO \\
            \hline
            \multirow{0.5}[4]{*}{DTLZ1} & 2     & 8     & \textbf{2.7212e+1 (1.08e+1) $\approx$} & 3.0662e+1 (8.78e+0) \\
            \cline{2-5}      & 3     & 6     & 9.8401e+0 (4.89e+0) $\approx$ & \textbf{9.4145e+0 (4.68e+0)} \\
            \hline
            \multirow{0.5}[4]{*}{DTLZ2} & 2     & 8     & 4.1878e-3 (9.77e-4) - & \textbf{3.4063e-3 (4.00e-4)} \\
            \cline{2-5}      & 3     & 6     & 2.2805e-2 (1.53e-3) - & \textbf{2.2150e-2 (1.83e-3)} \\
            \hline
            \multirow{0.5}[4]{*}{DTLZ3} & 2     & 8     & \textbf{6.9145e+1 (2.99e+1) $\approx$} & 7.6125e+1 (2.95e+1) \\
            \cline{2-5}      & 3     & 6     & \textbf{3.8567e+1 (1.80e+1) $\approx$} & 4.0630e+1 (1.72e+1) \\
            \hline
            \multirow{0.5}[4]{*}{DTLZ4} & 2     & 8     & \textbf{1.6091e-1 (1.19e-1) $\approx$} & 1.9344e-1 (1.35e-1) \\
            \cline{2-5}      & 3     & 6     & 1.0577e-1 (4.72e-2) $\approx$ & \textbf{9.2780e-2 (3.91e-2)} \\
            \hline
            \multirow{0.5}[4]{*}{DTLZ5} & 2     & 8     & 3.7814e-3 (7.96e-4) $\approx$ & \textbf{3.4790e-3 (4.10e-4)} \\
            \cline{2-5}      & 3     & 6     & 5.2674e-3 (9.79e-4) - & \textbf{2.9033e-3 (1.68e-4)} \\
            \hline
            \multirow{0.5}[4]{*}{DTLZ6} & 2     & 8     & 1.1725e+0 (3.14e-1) - & \textbf{8.8361e-1 (3.18e-1)} \\
            \cline{2-5}      & 3     & 6     & 3.9026e-1 (9.49e-2) - & \textbf{3.2192e-1 (1.67e-1)} \\
            \hline
            \multirow{0.5}[4]{*}{DTLZ7} & 2     & 8     & 6.9455e-3 (1.16e-3) - & \textbf{1.9779e-3 (6.93e-5)} \\
            \cline{2-5}      & 3     & 6     & 3.9822e-2 (1.76e-3) - & \textbf{2.2759e-2 (6.08e-4)} \\
            \hline
            ZDT1  & 2     & 8     & 1.1698e-2 (3.50e-3) - & \textbf{2.9639e-3 (3.48e-4)} \\
            \hline
            ZDT2  & 2     & 8     & 5.5068e-3 (7.13e-4) - & \textbf{2.6829e-3 (1.33e-4)} \\
            \hline
            ZDT3  & 2     & 8     & 1.6370e-2 (1.42e-2) - & \textbf{2.6870e-3 (5.41e-4)} \\
            \hline
            ZDT4  & 2     & 8     & 2.5442e+1 (1.21e+1) $\approx$ & \textbf{2.2829e+1 (1.14e+1)} \\
            \hline
            ZDT6  & 2     & 8     & 6.5736e-1 (2.48e-1) - & \textbf{4.1481e-1 (1.10e-1)} \\
            \hline
            \multicolumn{3}{|c|}{+/-/$\approx$} & 0/11/8 &  \\
            \hline\hline
            \end{tabular}%
            }
            \label{tab:CI-EMO-no-Norm}%
        \end{table}%

        \subsection{Effectiveness of normalization}\label{Sec_IV_C}
        \par For integrating the three kinds of performance metrics of convergence, diversity, and distribution, the algorithm normalizes different indicators and then superimposes them together. To verify the impact of normalization on each indicator, we used a new variant CI-EMO-no-Norm as a control. In CI-EMO-no-Norm, $\mathbf{I}_1$, $\mathbf{I}_2$, and $\mathbf{I}_3$ are obtained without normalization, that is, they are equal to $ \theta(\boldsymbol{x}) $,  $ d_{c}\left(\boldsymbol{x}\right) $, and $ -d_{z}\left(\boldsymbol{x}\right) $, respectively. The experimental result is shown in Table~\ref{tab:CI-EMO-no-Norm}. CI-EMO significantly outperformed CI-EMO-no-Norm on 11 test questions, and there was no significant difference on the remaining 8 test questions. It shows that normalization can help combine different scale metrics and achieve the balance of convergence, diversity, and distribution.

         \begin{table*}[h]\scriptsize
		  \centering
		  \caption{\centering IGD+ STATISTIC RESULTS (MEAN AND STANDARD DEVIATION) FOR CI-EMO AND OTHER FOUR VARIANTS WITH DIFFERENT SAMPLING NUMBER}
          \resizebox{0.9\linewidth}{!}{
            \begin{tabular}{|c|c|c|c|c|c|c|c|}
            \hline\hline
            Problem & M     & D     & CI-EMO-q10 & CI-EMO-q5 & CI-EMO-q3 & CI-EMO-q2 & CI-EMO \\
            \hline
            \multirow{0.5}[4]{*}{DTLZ1} & 2     & 8     & 3.7446e+1 (1.28e+1) $\approx$ & \textbf{2.9896e+1 (1.25e+1) $\approx$} & 3.1577e+1 (1.02e+1) $\approx$ & 3.2388e+1 (1.67e+1) $\approx$ & 3.0662e+1 (8.78e+0) \\
            \cline{2-8}      & 3     & 6     & 1.1769e+1 (5.36e+0) $\approx$ & 1.2843e+1 (5.59e+0) - & 1.3591e+1 (5.49e+0) - & 9.8632e+0 (5.32e+0) $\approx$ & \textbf{9.4145e+0 (4.68e+0)} \\
            \hline
            \multirow{0.5}[4]{*}{DTLZ2} & 2     & 8     & 3.5973e-3 (4.40e-4) $\approx$ & 4.3215e-3 (2.74e-3) $\approx$ & 3.9059e-3 (1.19e-3) $\approx$ & \textbf{3.3842e-3 (5.53e-4) $\approx$} & 3.4063e-3 (4.00e-4) \\
            \cline{2-8}      & 3     & 6     & 2.2807e-2 (1.11e-3) - & 2.2577e-2 (9.52e-4) - & 2.2198e-2 (1.24e-3) $\approx$ & \textbf{2.1866e-2 (8.19e-4) $\approx$} & 2.2150e-2 (1.83e-3) \\
            \hline
            \multirow{0.5}[4]{*}{DTLZ3} & 2     & 8     & 8.9218e+1 (3.23e+1) $\approx$ & 8.1997e+1 (2.97e+1) $\approx$ & 8.2436e+1 (3.04e+1) $\approx$ & \textbf{7.6044e+1 (2.27e+1) $\approx$} & 7.6125e+1 (2.95e+1) \\
            \cline{2-8}      & 3     & 6     & 4.1273e+1 (1.79e+1) $\approx$ & \textbf{3.6468e+1 (1.28e+1) $\approx$} & 3.7524e+1 (1.64e+1) $\approx$ & 3.8996e+1 (2.18e+1) $\approx$ & 4.0630e+1 (1.72e+1) \\
            \hline
            \multirow{0.5}[4]{*}{DTLZ4} & 2     & 8     & 2.5848e-1 (1.18e-1) $\approx$ & 1.9311e-1 (1.26e-1) $\approx$ & 2.3712e-1 (1.32e-1) $\approx$ & \textbf{1.5557e-1 (1.17e-1) $\approx$} & 1.9344e-1 (1.35e-1) \\
            \cline{2-8}      & 3     & 6     & 1.0845e-1 (3.33e-2) $\approx$ & 1.2143e-1 (4.70e-2) - & 1.1937e-1 (6.47e-2) $\approx$ & 1.1902e-1 (5.64e-2) $\approx$ & \textbf{9.2780e-2 (3.91e-2)} \\
            \hline
            \multirow{0.5}[4]{*}{DTLZ5} & 2     & 8     & 3.6239e-3 (4.53e-4) $\approx$ & 3.6673e-3 (1.17e-3) $\approx$ & 3.6255e-3 (6.64e-4) $\approx$ & 3.7440e-3 (9.87e-4) $\approx$ & \textbf{3.4790e-3 (4.10e-4)} \\
            \cline{2-8}      & 3     & 6     & 2.8387e-3 (1.99e-4) $\approx$ & \textbf{2.8341e-3 (2.48e-4) $\approx$} & 2.8383e-3 (2.14e-4) $\approx$ & 2.9206e-3 (2.38e-4) $\approx$ & 2.9033e-3 (1.68e-4) \\
            \hline
            \multirow{0.5}[4]{*}{DTLZ6} & 2     & 8     & 1.2990e+0 (5.08e-1) - & 1.2013e+0 (4.32e-1) - & 1.1234e+0 (3.79e-1) - & 1.1415e+0 (4.09e-1) - & \textbf{8.8361e-1 (3.18e-1)} \\
            \cline{2-8}      & 3     & 6     & 4.8597e-1 (1.72e-1) - & 4.2108e-1 (1.63e-1) $\approx$ & \textbf{3.0554e-1 (9.43e-2) $\approx$} & 3.7405e-1 (1.22e-1) $\approx$ & 3.2192e-1 (1.67e-1) \\
            \hline
            \multirow{0.5}[4]{*}{DTLZ7} & 2     & 8     & 2.1768e-3 (1.68e-4) - & 2.1098e-3 (9.95e-5) - & 2.0516e-3 (9.56e-5) - & 2.0400e-3 (1.54e-4) $\approx$ & \textbf{1.9779e-3 (6.93e-5)} \\
            \cline{2-8}      & 3     & 6     & 2.3407e-2 (8.60e-4) - & 2.3042e-2 (7.98e-4) $\approx$ & 2.2876e-2 (6.90e-4) $\approx$ & 2.2971e-2 (7.42e-4) $\approx$ & \textbf{2.2759e-2 (6.08e-4)} \\
            \hline
            ZDT1  & 2     & 8     & 3.1478e-3 (5.12e-4) $\approx$ & 3.1382e-3 (3.67e-4) $\approx$ & 3.0683e-3 (4.22e-4) $\approx$ & 3.1799e-3 (5.57e-4) $\approx$ & \textbf{2.9639e-3 (3.48e-4)} \\
            \hline
            ZDT2  & 2     & 8     & 2.9333e-3 (2.16e-4) - & 2.7666e-3 (1.77e-4) $\approx$ & 2.6943e-3 (1.57e-4) $\approx$ & 2.7713e-3 (1.65e-4) $\approx$ & \textbf{2.6829e-3 (1.33e-4)} \\
            \hline
            ZDT3  & 2     & 8     & \textbf{2.2085e-3 (2.85e-4) +} & 2.4203e-3 (3.56e-4) $\approx$ & 2.4877e-3 (3.99e-4) $\approx$ & 2.4821e-3 (4.74e-4) $\approx$ & 2.6870e-3 (5.41e-4) \\
            \hline
            ZDT4  & 2     & 8     & 2.2918e+1 (1.04e+1) $\approx$ & \textbf{1.9470e+1 (9.91e+0) $\approx$} & 2.3751e+1 (1.09e+1) $\approx$ & 2.2798e+1 (1.03e+1) $\approx$ & 2.2829e+1 (1.14e+1) \\
            \hline
            ZDT6  & 2     & 8     & 8.7485e-1 (7.23e-1) - & 5.4028e-1 (2.28e-1) $\approx$ & 4.6791e-1 (1.34e-1) $\approx$ & 5.7742e-1 (2.32e-1) - & \textbf{4.1481e-1 (1.10e-1)} \\
            \hline
            \multicolumn{3}{|c|}{+/-/$\approx$} & 1/7/11 & 0/5/14 & 0/3/16 & 0/2/17 &  \\
            \hline\hline
            \end{tabular}%
            }
            \label{tab:CI-EMO-num}%
        \end{table*}%

        \subsection{Effect of random weights of indicators}\label{Sec_IV_C}
        \par
\begin{figure}[h]
    		\centering
    		\resizebox{0.7\linewidth}{!}{
    			\includegraphics[scale=1]{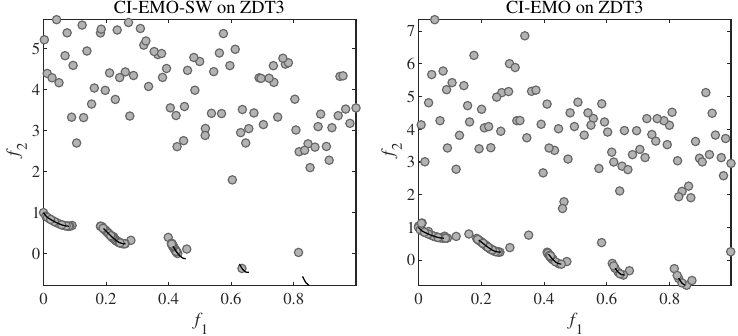}
    		}
    		\caption{Evaluated solutions obtained by CI-EMO and CI-EMO-SW on ZDT3.}
    		\label{Figure_SW}
	\end{figure}

 A fixed weight setting may cause the algorithm's sampling points to exhibit a single preference and lack diversity. To address this, random weights are assigned to the three sampling indicators during their integration. To analyze the impact of random weights, we compare CI-EMO with its variant, CI-EMO-SW, where all weights are fixed and set to 1. The results are presented in Table~S-III. Experimental comparisons reveal that the overall statistical results of fixed weights and random weights are not significantly different. However, random weights yield more optimal results, as highlighted in Table~S-III. Furthermore, as shown in Fig.~\ref{Figure_SW}, for the discontinuous Pareto front problem ZDT3, using random weights demonstrates better robustness in identifying each segment of the discontinuous Pareto front. Experiments show random weights enhance, improve the diversity of sampling, and prevent issues due to fixed weights.

        \subsection{Effect of sampling number setting each generation}\label{Sec_IV_C}
        \par In expensive optimization, the sampling number in each iteration is commonly an important parameter for algorithm performance. Some previous methods use multiple sampling strategies to select multiple candidate solutions from the candidate population at each iteration for real fitness evaluation, and each strategy is used for different purposes~\citep{Chugh2016-KRVEA, Song2021-KTA2}. In CI-EMO, we use a composite indicator to select only one candidate in each iteration, where the selected candidate balances the performance of convergence, diversity, and distribution.

        \par To verify the impact of the sampling number of CI-EMO in this paper, we compared CI-EMO with four variants, where CI-EMO-q10 means the algorithm selects 10 candidate solutions for real fitness evaluation in each iteration. Similarly, CI-EMO-q5, CI-EMO-q3, and CI-EMO-q2 respectively indicate that 5, 3, and 2 candidate solutions are selected from the candidate population for real fitness value evaluation in each iteration. When the former candidate solution is selected based on the composite indicator, the candidate solution is deleted from the candidate population, and the method selects the next new candidate solution. As shown in Table~\ref{tab:CI-EMO-num}, CI-EMO has the best performance. The performance of CI-EMO-q2 and CI-EMO-q3 are similar to CI-EMO. CI-EMO-q3 performs significantly worse than CI-EMO on only 3 test problems. As the number of samples continues to increase, the performance of the algorithm decreases significantly. This is because some unnecessary candidate solutions in the population are selected, making the use of the true fitness value evaluation times no longer efficient.

        \subsection{Computational Complexity Analysis}\label{Sec_IV_F}
        \par In each iteration, the computational overhead of CI-EMO can be categorized into three main components: training the Gaussian Process model, surrogate-assisted NSGA-III search candidate population, and selecting one query point from the candidate population. 1) Training the GP Model: The computational complexity of training the GP model for all objectives is $O\left(m n^3\right)$, where $n$ denotes the number of real fitness evaluations and $m$ represents the number of objectives.
        2) Surrogate-assisted NSGA-III: The computational complexity for generating the candidate population includes the NSGA-III operations and the time required for the GP model to predict objective values. The computational complexity of NSGA-III operations is $O\left(G m N^2\right)$, where $N$ is the population size and $G$ is the number of iterations. The computational complexity of GP model to predict objective values is $O\left(G N*m n^2\right)$, where $GN$ is the times using surrogate model to predict fitness value and $n$ is the sample number of training model. Due to $n$ commonly is larger than $N$, so total the computational complexity is $O\left(G N*m n^2\right)$.
        3) Selecting candidate point: The computational complexity of composite indicator consists of three indicators calculation. The computational complexity of $I_1$, $I_2$, and $I_3$ are $O(mn_{nd}N)$, $O(mnN)$, and $O(mN)$, respectively. So the computational complexity of the composite indicator is $O(mnN)$. In a word, the computational complexity of selecting the candidate point can be ignored compared to the other two processes (training the GP Model and surrogate-assisted NSGA-III), and the total complexity in each iteration of CI-EMO is $O(m n^3 + G N m n^2)$.

        \par
\begin{figure}[ht]
    		\centering
    		\resizebox{0.7\linewidth}{!}{
    			\includegraphics[scale=1]{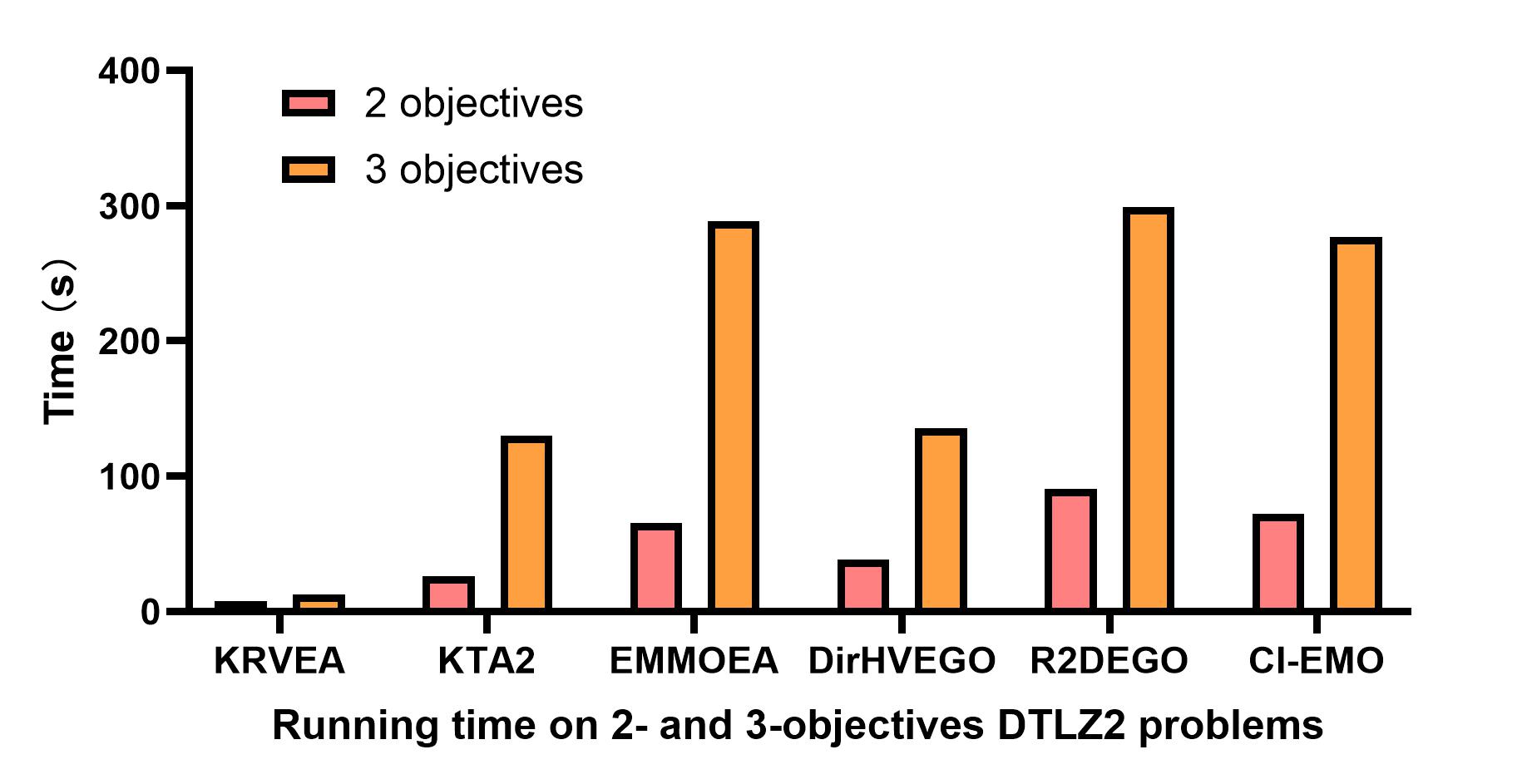}
    		}
    		\caption{Comparison of running time between CI-EMO and five SAEAs.}
    		\label{Figure_time}
	\end{figure}

 In addition, we compared the running time of CI-EMO with five SAEAs on DTLZ2 problems with 2 and 3 objectives, as shown in Figure~\ref{Figure_time}. Due to CI-EMO samples only one candidate in each iteration, its running time is relatively high. But the running time of CI-EMO is less than EMMOEA and R2DEGO on 3- objective DTLZ2 problem, where R2DEGO samples 5 samples in each iteration. Overall, the running time of CI-EMO is comparable to other SAEAs and is reasonable for expensive optimization scenarios since expensive fitness evaluation is time-consuming, for example, once computational fluid dynamics simulations could take hours~\citep{Song2021-KTA2}.

        \begin{table*}[ht]\scriptsize
		  \centering
		  \caption{\centering HV STATISTIC RESULTS (MEAN AND STANDARD DEVIATION) FOR CI-EMO AND OTHER FIVE SAEAS ALGORITHMS ON THREE REAL-WORLD PROBLEMS}
            \resizebox{\linewidth}{!}{
            \begin{tabular}{|ccc|c|c|c|c|c|c|}
            \hline\hline
            \multicolumn{1}{|c|}{Problem} & \multicolumn{1}{c|}{M} & D     & K-RVEA & KTA2  & EMMOEA & DirHV-EGO & R2/D-EGO & CI-EMO \\
            \hline
            \multicolumn{1}{|c|}{Gear train design problem} & \multicolumn{1}{c|}{2} & 4     & 4.6450e-1 (4.87e-3) - & 4.7288e-1 (2.35e-3) - & 4.5895e-1 (9.15e-3) - & 4.6983e-1 (3.78e-3) - & 4.6625e-1 (5.30e-3) - & \textbf{4.7816e-1 (1.67e-3)} \\
            \hline
            \multicolumn{1}{|c|}{Car side impact design problem} & \multicolumn{1}{c|}{3} & 7     & 2.3479e-2 (3.44e-4) $\approx$ & 2.3794e-2 (2.58e-4) $\approx$ & 2.2190e-2 (5.42e-4) - & 2.3524e-2 (3.00e-4) $\approx$ & \textbf{2.3950e-2 (3.63e-4) +} & 2.3686e-2 (3.58e-4) \\
            \hline
            \multicolumn{1}{|c|}{Two bar plane truss} & \multicolumn{1}{c|}{2} & 2     & 8.4443e-1 (1.49e-3) - & 8.3838e-1 (6.46e-3) - & 8.4608e-1 (7.91e-4) - & 8.4696e-1 (3.94e-4) $\approx$ & 8.4664e-1 (3.15e-4) - & \textbf{8.4700e-1 (2.88e-4)} \\
            \hline
            \multicolumn{3}{|c|}{+/-/$\approx$} & 0/2/1 & 0/2/1 & 0/3/0 & 0/1/2 & 1/2/0 &  \\
            \hline\hline
            \end{tabular}%
            }
            \label{tab:SAEAs-RW-HV}%
        \end{table*}%
        
        \subsection{Comparison with state-of-the-art SAEAs on real-world problems}\label{Sec_IV_E}
        \par In this section, we empirically evaluate our proposed method based on three real-world MOPs: gear train design problem~\citep{Ray2002-swarm}, car side impact design problem~\citep{Jain2013-evolutionary}, and two bar plane truss~\cite{Coello2005-multiobjective}, the detailed description of problems can be seen in the references due to space constraints. Their objective and decision variable numbers are shown in Table~\ref{tab:SAEAs-RW-HV}. We compared our approach with five advanced SAEAs. For each algorithm, the number of initial samples $\mathcal{N}_0$ is set to 100, other experimental setup was consistent with our previous study, and all experiments are conducted over 21 independent runs to ensure statistical reliability. To summarize the performance, we recorded the hypervolume values obtained for each method, which are reported in Table~\ref{tab:SAEAs-RW-HV}.

        \par Based on the experimental results, CI-EMO clearly achieves the highest hypervolume scores for the gear train design problem and the two-bar plane truss problem, outperforming five advanced SAEAs and showcasing its superior performance. Specifically, CI-EMO outperforms EMMOEA across all three real-world MOPs in terms of hypervolume values. Moreover, CI-EMO secures the third-best results in the car side impact design problem. Fig.~S-2 shows the nondominated solutions obtained from six algorithms under comparison on three real-world problems. CI-EMO performs much better in improving the distributivity of the Pareto solution set compared to other state-of-the-art methods. It is further highlighting its effectiveness and competitive advantage in solving real-world expensive multi-objective optimization problems.

        \section{CONCLUSION}\label{Sec_V}
        \par In this work, we propose a composite indicator-guided infilling sampling for expensive multi-objective optimization. This method integrates three sampling metrics into a composite indicator. The developed sampling strategy effectively balances convergence, diversity, and distribution, while simplifying algorithm design through a modular candidate selection. Experimental results demonstrate that the composite indicator effectively balances the performance of convergence, diversity, and distribution, enabling CI-EMO to solve expensive multi-objective optimization problems effectively. Notably, the set of Pareto solutions obtained by our method significantly outperforms existing advanced methods in terms of distributivity. Additionally, the composite indicator is computationally inexpensive, with its computational complexity being negligible compared to training the GP model and surrogate-assisted NSGA-III.
        \par Despite these contributions, this study has some limitations. The random setting of indicator weights, while increasing robustness, has resulted in a loss of specificity. In future research, we plan to investigate adaptive weight adjustment strategies to address these limitations and extend the composite indicator-based EMO approach to more optimization scenarios, such as large-scale optimization problems.

        \section*{Acknowledgment}
        This work was partly supported by the National Natural Science Foundation of China under Grant Nos. 62403442 and 42327803, the Postdoctoral Fellowship Program of CPSF under Grant No. GZC20241600, the "CUG Scholar" Scientific Research Funds at China University of Geosciences (Wuhan) (Project No. 2024011), the Fundamental Research Founds for National University, China University of Geosciences (Wuhan), as well as support from the Hubei Province Postdoctoral Talent Introduction Program (Project No. 2024HBBHJD096).
    
	\bibliographystyle{elsarticle-harv}
	\bibliography{references}

    \newpage
	\onecolumn
	\setcounter{page}{1}
	
	\section*{\Large Supplementary Files for ``Composite Indicator-Guided Infilling Sampling for Expensive Multi-Objective Optimization''}
	
	\setcounter{table}{0}
	\renewcommand\thetable{S-\Roman{table}}
	
	\setcounter{figure}{0}
	\renewcommand\thefigure{S-\arabic{figure}}

        \begin{table*}[ht]\footnotesize
		  \centering
		  \caption{\centering HV RESULTS (MEAN AND STANDARD DEVIATION) OBTAINED BY THE PROPOSED CI-EMO AND OTHER FIVE SAEAS ALGORITHMS ON DTLZ AND ZDT BENCHMARK PROBLEMS}
            \resizebox{\linewidth}{!}{
            \begin{tabular}{|c|c|c|c|c|c|c|c|c|}
            \hline
            Problem & M     & D     & K-RVEA & KTA2  & EMMOEA & DirHV-EGO & R2/D-EGO & CI-EMO \\
            \hline
            \multirow{0.5}[4]{*}{DTLZ1} & 2     & 8     & 0.0000e+0 (0.00e+0) $\approx$ & 0.0000e+0 (0.00e+0) $\approx$ & 0.0000e+0 (0.00e+0) $\approx$ & 0.0000e+0 (0.00e+0) $\approx$ & 0.0000e+0 (0.00e+0) $\approx$ & 0.0000e+0 (0.00e+0) \\
            \cline{2-9}      & 3     & 6     & 0.0000e+0 (0.00e+0) $\approx$ & 0.0000e+0 (0.00e+0) $\approx$ & 0.0000e+0 (0.00e+0) $\approx$ & 0.0000e+0 (0.00e+0) $\approx$ & 0.0000e+0 (0.00e+0) $\approx$ & 0.0000e+0 (0.00e+0) \\
            \hline
            \multirow{0.5}[4]{*}{DTLZ2} & 2     & 8     & 3.0461e-1 (2.70e-2) - & 3.4051e-1 (1.92e-3) - & 3.3749e-1 (2.87e-3) - & 3.4073e-1 (2.68e-3) - & 3.3874e-1 (1.26e-3) - & \textbf{3.4449e-1 (8.35e-4)} \\
            \cline{2-9}      & 3     & 6     & 5.3657e-1 (3.85e-3) - & 5.6209e-1 (2.55e-3) - & 5.6248e-1 (3.21e-3) $\approx$ & 5.5908e-1 (2.40e-3) - & 5.4462e-1 (2.22e-3) - & \textbf{5.6402e-1 (3.22e-3)} \\
            \hline
            \multirow{0.5}[4]{*}{DTLZ3} & 2     & 8     & 0.0000e+0 (0.00e+0) $\approx$ & 0.0000e+0 (0.00e+0) $\approx$ & 0.0000e+0 (0.00e+0) $\approx$ & 0.0000e+0 (0.00e+0) $\approx$ & 0.0000e+0 (0.00e+0) $\approx$ & 0.0000e+0 (0.00e+0) \\
            \cline{2-9}      & 3     & 6     & 0.0000e+0 (0.00e+0) $\approx$ & 0.0000e+0 (0.00e+0) $\approx$ & 0.0000e+0 (0.00e+0) $\approx$ & 0.0000e+0 (0.00e+0) $\approx$ & 0.0000e+0 (0.00e+0) $\approx$ & 0.0000e+0 (0.00e+0) \\
            \hline
            \multirow{0.5}[4]{*}{DTLZ4} & 2     & 8     & 8.3885e-2 (3.05e-2) - & 1.3377e-1 (7.60e-2) $\approx$ & \textbf{1.8070e-1 (7.24e-2) $\approx$} & 8.5441e-2 (4.27e-2) - & 8.5301e-2 (6.69e-2) - & 1.4548e-1 (9.17e-2) \\
            \cline{2-9}      & 3     & 6     & 3.2600e-1 (1.04e-1) - & 4.1129e-1 (8.52e-2) $\approx$ & 4.0491e-1 (7.40e-2) $\approx$ & 1.7279e-1 (7.68e-2) - & 1.5465e-1 (6.93e-2) - & \textbf{4.2087e-1 (8.05e-2)} \\
            \hline
            \multirow{0.5}[4]{*}{DTLZ5} & 2     & 8     & 3.1256e-1 (7.70e-3) - & 3.4147e-1 (1.56e-3) - & 3.3611e-1 (3.53e-3) - & 3.4112e-1 (1.06e-3) - & 3.3850e-1 (1.24e-3) - & \textbf{3.4430e-1 (8.24e-4)} \\
            \cline{2-9}      & 3     & 6     & 1.8349e-1 (2.62e-3) - & \textbf{1.9971e-1 (2.04e-4) +} & 1.9668e-1 (1.23e-3) - & 1.9484e-1 (4.21e-4) - & 1.9874e-1 (2.29e-4) - & 1.9929e-1 (2.27e-4) \\
            \hline
            \multirow{0.5}[4]{*}{DTLZ6} & 2     & 8     & 0.0000e+0 (0.00e+0) $\approx$ & 4.3290e-3 (1.98e-2) $\approx$ & 0.0000e+0 (0.00e+0) $\approx$ & \textbf{2.0703e-1 (9.48e-2) +} & 1.6227e-1 (9.84e-2) + & 8.6580e-3 (2.73e-2) \\
            \cline{2-9}      & 3     & 6     & 3.8080e-3 (1.30e-2) - & 5.3468e-2 (4.75e-2) $\approx$ & 5.6954e-2 (4.16e-2) $\approx$ & 1.3765e-1 (2.53e-2) + & \textbf{1.7115e-1 (2.61e-2) +} & 5.3269e-2 (4.49e-2) \\
            \hline
            \multirow{0.5}[4]{*}{DTLZ7} & 2     & 8     & 2.3343e-1 (2.57e-3) - & 2.3488e-1 (1.97e-2) - & 2.3793e-1 (2.35e-3) - & 2.1938e-1 (1.91e-2) - & 2.3006e-1 (6.73e-3) - & \textbf{2.4279e-1 (4.80e-5)} \\
            \cline{2-9}      & 3     & 6     & 2.6756e-1 (1.77e-3) - & 2.6290e-1 (2.65e-2) - & 2.6140e-1 (2.14e-2) - & 2.7680e-1 (4.06e-3) - & 2.7841e-1 (1.01e-3) - & \textbf{2.8270e-1 (3.83e-4)} \\
            \hline
            ZDT1  & 2     & 8     & 6.9423e-1 (3.91e-3) - & 7.1298e-1 (3.42e-3) - & 6.8293e-1 (2.74e-2) - & 7.1799e-1 (5.39e-4) - & 7.1905e-1 (1.41e-3) $\approx$ & \textbf{7.1931e-1 (4.28e-4)} \\
            \hline
            ZDT2  & 2     & 8     & 4.0796e-1 (2.62e-2) - & 4.3992e-1 (1.16e-3) - & 4.3718e-1 (9.65e-3) - & 4.4345e-1 (3.93e-4) - & 4.4280e-1 (8.90e-4) - & \textbf{4.4416e-1 (2.88e-4)} \\
            \hline
            ZDT3  & 2     & 8     & 5.8451e-1 (2.15e-2) - & 6.2750e-1 (7.88e-2) $\approx$ & \textbf{6.3215e-1 (1.09e-1) $\approx$} & 6.2017e-1 (4.58e-2) $\approx$ & 5.9710e-1 (8.82e-4) - & 6.0010e-1 (2.55e-3) \\
            \hline
            ZDT4  & 2     & 8     & 0.0000e+0 (0.00e+0) $\approx$ & 0.0000e+0 (0.00e+0) $\approx$ & 0.0000e+0 (0.00e+0) $\approx$ & 0.0000e+0 (0.00e+0) $\approx$ & 0.0000e+0 (0.00e+0) $\approx$ & 0.0000e+0 (0.00e+0) \\
            \hline
            ZDT6  & 2     & 8     & 3.7319e-5 (1.71e-4) - & 4.5315e-2 (5.79e-2) $\approx$ & 5.7842e-2 (6.72e-2) $\approx$ & 2.2210e-1 (7.99e-2) + & \textbf{2.2857e-1 (8.38e-2) +} & 4.3572e-2 (3.57e-2) \\
            \hline
            \multicolumn{3}{|c|}{+/-/$\approx$} & 0/13/6 & 1/7/11 & 0/7/12 & 3/10/6 & 3/10/6 &  \\
            \hline
            \end{tabular}%
            }
            \label{tab:SAEAs-HV}%
        \end{table*}%

        \begin{table*}[ht]\footnotesize
		  \centering
		  \caption{\centering HV RESULTS (MEAN AND STANDARD DEVIATION) OBTAINED BY THE PROPOSED CI-EMO AND OTHER FIVE SAEAS ALGORITHMS ON MaF BENCHMARK PROBLEMS}
            \resizebox{\linewidth}{!}{
            \begin{tabular}{|c|c|c|c|c|c|c|c|c|}
            \hline
            Problem & M     & D     & KRVEA & KTA2  & EMMOEA & DirHVEGO & R2DEGO & CI-EMO \\
            \hline
            \multirow{1.5}[6]{*}{MaF1} & 3     & 10    & 1.9027e-1 (3.99e-3) - & 2.1218e-1 (6.78e-3) - & 2.1419e-1 (2.26e-3) - & 2.0125e-1 (2.85e-3) - & 1.9899e-1 (3.23e-3) - & \textbf{2.2107e-1 (1.82e-3)} \\
            \cline{2-9}      & 5     & 10    & 8.5585e-3 (8.71e-4) - & 7.9485e-3 (7.68e-4) - & 1.0061e-2 (5.99e-4) $\approx$ & 6.4275e-3 (2.50e-4) - & 8.0139e-3 (2.94e-4) - & \textbf{1.0316e-2 (3.64e-4)} \\
            \cline{2-9}      & 10    & 10    & 1.0202e-7 (2.06e-8) - & 2.1497e-7 (9.91e-8) - & 5.1605e-7 (7.58e-8) - & 1.0458e-7 (2.10e-8) - & 1.1104e-7 (1.97e-7) - & \textbf{7.2391e-7 (1.27e-7)} \\
            \hline
            \multirow{1.5}[6]{*}{MaF2} & 3     & 10    & 2.3310e-1 (4.62e-3) - & 2.4280e-1 (1.30e-3) - & 2.3329e-1 (3.76e-3) - & \textbf{2.4737e-1 (3.67e-4) +} & 2.4169e-1 (1.20e-3) - & 2.4492e-1 (1.27e-3) \\
            \cline{2-9}      & 5     & 10    & 1.8142e-1 (3.24e-3) - & 1.9020e-1 (2.02e-3) - & 1.8425e-1 (3.26e-3) - & \textbf{1.9658e-1 (1.28e-3) $\approx$} & 1.9235e-1 (2.23e-3) - & 1.9576e-1 (2.68e-3) \\
            \cline{2-9}      & 10    & 10    & 2.1822e-1 (3.39e-3) - & 2.0633e-1 (2.29e-3) - & 2.1672e-1 (4.26e-3) - & 2.1638e-1 (2.38e-3) - & \textbf{2.2898e-1 (1.58e-3) +} & 2.2122e-1 (2.25e-3) \\
            \hline
            \multirow{1.5}[6]{*}{MaF3} & 3     & 10    & 0.0000e+0 (0.00e+0) $\approx$ & 0.0000e+0 (0.00e+0) $\approx$ & 0.0000e+0 (0.00e+0) $\approx$ & 0.0000e+0 (0.00e+0) $\approx$ & 0.0000e+0 (0.00e+0) $\approx$ & 0.0000e+0 (0.00e+0) \\
            \cline{2-9}      & 5     & 10    & 0.0000e+0 (0.00e+0) $\approx$ & 0.0000e+0 (0.00e+0) $\approx$ & 0.0000e+0 (0.00e+0) $\approx$ & 0.0000e+0 (0.00e+0) $\approx$ & 0.0000e+0 (0.00e+0) $\approx$ & 0.0000e+0 (0.00e+0) \\
            \cline{2-9}      & 10    & 10    & 5.0320e-2 (1.11e-1) $\approx$ & \textbf{6.1107e-1 (3.42e-1) +} & 1.3805e-1 (1.95e-1) $\approx$ & 2.4563e-1 (2.49e-1) + & 3.3580e-1 (3.05e-1) + & 9.9335e-2 (2.48e-1) \\
            \hline
            \multirow{1.5}[6]{*}{MaF4} & 3     & 10    & 0.0000e+0 (0.00e+0) $\approx$ & 0.0000e+0 (0.00e+0) $\approx$ & 0.0000e+0 (0.00e+0) $\approx$ & 0.0000e+0 (0.00e+0) $\approx$ & 0.0000e+0 (0.00e+0) $\approx$ & 0.0000e+0 (0.00e+0) \\
            \cline{2-9}      & 5     & 10    & 0.0000e+0 (0.00e+0) $\approx$ & 0.0000e+0 (0.00e+0) $\approx$ & 0.0000e+0 (0.00e+0) $\approx$ & 0.0000e+0 (0.00e+0) $\approx$ & 0.0000e+0 (0.00e+0) $\approx$ & 0.0000e+0 (0.00e+0) \\
            \cline{2-9}      & 10    & 10    & 1.1402e-7 (1.83e-7) - & 1.7436e-6 (2.39e-6) $\approx$ & \textbf{1.1127e-5 (1.66e-5) $\approx$} & 2.3326e-7 (4.86e-7) - & 6.9921e-7 (1.19e-6) - & 3.5638e-6 (4.31e-6) \\
            \hline
            \multirow{1.5}[6]{*}{MaF5} & 3     & 10    & 5.8546e-2 (5.51e-2) - & 2.4347e-1 (1.16e-1) $\approx$ & \textbf{2.7857e-1 (1.26e-1) +} & 5.3432e-2 (3.94e-2) - & 5.6590e-2 (4.69e-2) - & 1.6657e-1 (1.25e-1) \\
            \cline{2-9}      & 5     & 10    & 1.1540e-1 (7.24e-2) - & 2.2795e-1 (1.21e-1) $\approx$ & \textbf{3.6663e-1 (1.53e-1) +} & 7.0922e-2 (4.65e-2) - & 8.0312e-2 (4.85e-2) - & 1.9786e-1 (1.18e-1) \\
            \cline{2-9}      & 10    & 10    & 7.4476e-1 (4.87e-2) - & 7.6957e-1 (7.46e-2) - & \textbf{8.6964e-1 (5.86e-2) +} & 6.2825e-1 (7.80e-2) - & 7.4705e-1 (4.32e-2) - & 8.3492e-1 (5.85e-2) \\
            \hline
            \multirow{1.5}[6]{*}{MaF6} & 3     & 10    & 2.8335e-4 (1.30e-3) - & 1.0583e-1 (3.79e-2) - & \textbf{1.6642e-1 (1.04e-2) +} & 3.1905e-3 (8.83e-3) - & 1.0044e-2 (1.26e-2) - & 1.3200e-1 (2.81e-2) \\
            \cline{2-9}      & 5     & 10    & 4.7388e-3 (1.19e-2) - & 1.0321e-1 (1.15e-2) + & \textbf{1.1474e-1 (2.86e-3) +} & 1.4998e-2 (1.57e-2) $\approx$ & 2.6747e-2 (2.40e-2) $\approx$ & 1.6809e-2 (2.57e-2) \\
            \cline{2-9}      & 10    & 10    & 9.6658e-2 (1.43e-3) - & 9.8905e-2 (1.65e-3) - & 9.8202e-2 (9.16e-4) - & 9.3467e-2 (1.49e-3) - & 9.7198e-2 (1.03e-3) - & \textbf{9.9933e-2 (3.64e-4)} \\
            \hline
            \multirow{1.5}[6]{*}{MaF7} & 3     & 10    & 2.5209e-1 (3.70e-3) - & 2.5443e-1 (2.54e-2) - & 2.5836e-1 (2.44e-2) - & 2.6660e-1 (5.26e-3) - & 2.6983e-1 (3.14e-3) - & \textbf{2.8034e-1 (9.87e-4)} \\
            \cline{2-9}      & 5     & 10    & 2.1118e-1 (6.01e-3) - & 2.3450e-1 (1.16e-2) $\approx$ & 2.3073e-1 (1.15e-2) $\approx$ & \textbf{2.4569e-1 (6.20e-3) +} & 2.3236e-1 (6.93e-3) $\approx$ & 2.3450e-1 (4.81e-3) \\
            \cline{2-9}      & 10    & 10    & 1.7515e-1 (5.57e-3) - & 1.6996e-1 (1.20e-2) - & 1.7435e-1 (6.23e-3) - & 1.4747e-1 (8.54e-3) - & 1.4603e-1 (7.26e-3) - & \textbf{1.8201e-1 (8.98e-3)} \\
            \hline
            \multirow{1.5}[6]{*}{MaF10} & 3     & 10    & 1.4257e-1 (4.47e-2) $\approx$ & 1.5406e-1 (5.85e-2) $\approx$ & \textbf{1.5921e-1 (4.38e-2) $\approx$} & 2.9224e-2 (3.95e-2) - & 2.0331e-2 (2.76e-2) - & 1.3722e-1 (4.18e-2) \\
            \cline{2-9}      & 5     & 10    & 1.9544e-1 (3.39e-2) - & 1.9753e-1 (6.91e-2) $\approx$ & 1.9958e-1 (5.49e-2) $\approx$ & 1.2859e-1 (5.06e-2) - & 1.2046e-1 (3.56e-2) - & \textbf{2.2253e-1 (5.30e-2)} \\
            \cline{2-9}      & 10    & 10    & 2.2149e-1 (1.69e-2) - & 6.4517e-1 (2.65e-1) - & 6.2589e-1 (2.32e-1) - & 2.1032e-1 (6.87e-2) - & 2.2924e-1 (1.07e-2) - & \textbf{8.7050e-1 (9.36e-2)} \\
            \hline
            \multirow{1.5}[6]{*}{MaF11} & 3     & 10    & 7.4666e-1 (5.05e-2) - & 7.4308e-1 (3.43e-2) - & 8.2768e-1 (1.53e-2) - & 7.9595e-1 (1.76e-2) - & 8.0300e-1 (1.11e-2) - & \textbf{8.5546e-1 (1.47e-2)} \\
            \cline{2-9}      & 5     & 10    & 8.6388e-1 (4.10e-2) - & 8.3081e-1 (3.64e-2) - & 9.0993e-1 (4.04e-2) $\approx$ & 9.0046e-1 (1.37e-2) - & 9.0298e-1 (1.78e-2) - & \textbf{9.2302e-1 (2.40e-2)} \\
            \cline{2-9}      & 10    & 11    & 9.6832e-1 (1.09e-2) - & 9.7324e-1 (1.05e-2) - & 9.5402e-1 (4.88e-2) - & \textbf{9.9164e-1 (2.64e-3) $\approx$} & 9.7918e-1 (2.42e-2) - & 9.8989e-1 (4.82e-3) \\
            \hline
            \multirow{1.5}[6]{*}{MaF12} & 3     & 10    & 2.3963e-1 (2.90e-2) - & 2.8325e-1 (4.02e-2) $\approx$ & \textbf{3.4154e-1 (3.96e-2) +} & 2.8119e-1 (2.98e-2) $\approx$ & 2.9240e-1 (3.22e-2) $\approx$ & 2.9569e-1 (3.25e-2) \\
            \cline{2-9}      & 5     & 10    & 3.5976e-1 (3.97e-2) - & 4.3480e-1 (5.47e-2) $\approx$ & \textbf{5.0032e-1 (5.80e-2) +} & 4.2081e-1 (5.94e-2) $\approx$ & 4.1019e-1 (6.53e-2) $\approx$ & 4.4101e-1 (6.63e-2) \\
            \cline{2-9}      & 10    & 10    & 6.5057e-1 (5.00e-2) - & 7.4022e-1 (2.23e-2) $\approx$ & \textbf{7.6448e-1 (1.90e-2) $\approx$} & 5.9450e-1 (5.31e-2) - & 6.2427e-1 (4.95e-2) - & 7.4714e-1 (3.62e-2) \\
            \hline
            \multirow{1.5}[6]{*}{MaF13} & 3     & 10    & 1.6516e-1 (4.87e-2) - & 3.6088e-1 (4.20e-2) - & 3.9093e-1 (2.64e-2) $\approx$ & 4.4328e-1 (1.29e-2) + & \textbf{5.1229e-1 (4.68e-3) +} & 3.9014e-1 (2.20e-2) \\
            \cline{2-9}      & 5     & 10    & 0.0000e+0 (0.00e+0) - & 9.8391e-2 (3.32e-2) + & 6.0588e-2 (3.13e-2) + & \textbf{1.0549e-1 (2.63e-2) +} & 5.2907e-4 (1.50e-3) - & 1.7688e-2 (1.67e-2) \\
            \cline{2-9}      & 10    & 10    & 0.0000e+0 (0.00e+0) - & 1.2057e-2 (1.61e-2) + & 1.3949e-2 (1.61e-2) + & \textbf{3.9530e-2 (2.86e-2) +} & 0.0000e+0 (0.00e+0) - & 4.7121e-3 (8.03e-3) \\
            \hline
            \multicolumn{3}{|c|}{+/-/$\approx$} & 0/27/6 & 4/16/13 & 9/11/13 & 6/18/9 & 3/22/8 &  \\
            \hline
            \end{tabular}%
            }
            \label{tab:SAEAs-MaF-HV}%
        \end{table*}%

        \begin{figure*}[ht]
    		\centering
    		\resizebox{\linewidth}{!}{
    			\includegraphics[scale=0.27]{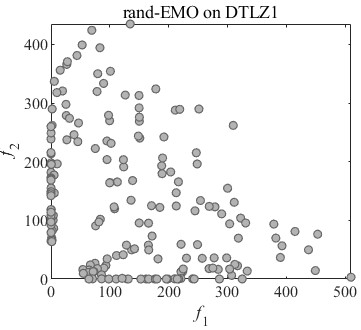}
                    \includegraphics[scale=0.27]{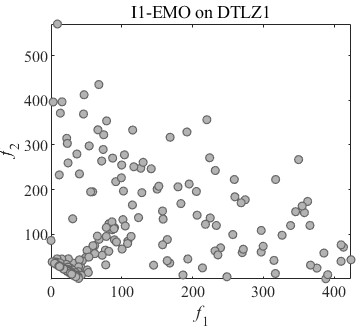}
                    \includegraphics[scale=0.27]{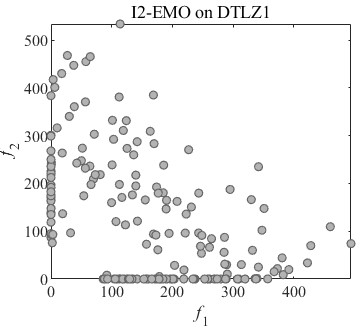}
                    \includegraphics[scale=0.27]{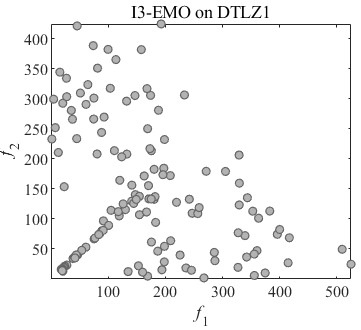}
                    \includegraphics[scale=0.27]{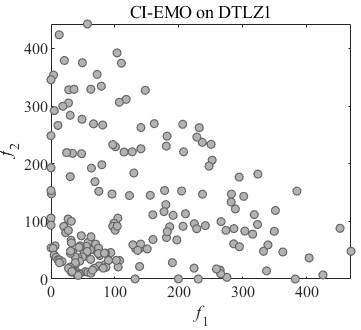}
                    }
                    
                \resizebox{\linewidth}{!}{
                    \includegraphics[scale=0.27]{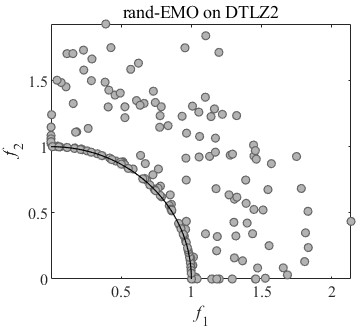}
                    \includegraphics[scale=0.27]{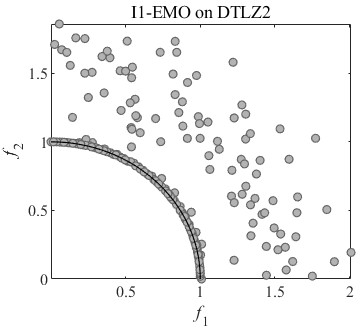}
                    \includegraphics[scale=0.27]{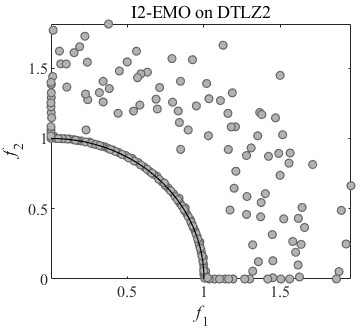}
                    \includegraphics[scale=0.27]{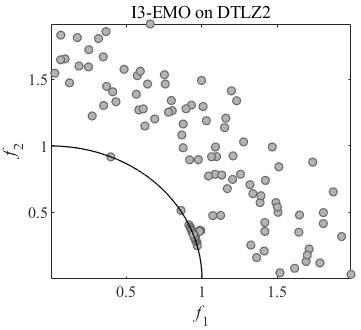}
                    \includegraphics[scale=0.27]{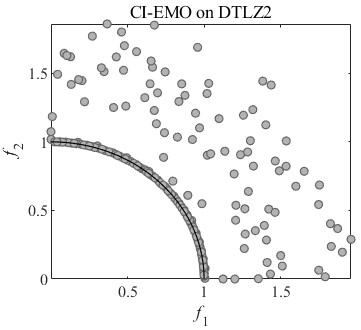}
                }

                \resizebox{\linewidth}{!}{
                    \includegraphics[scale=0.27]{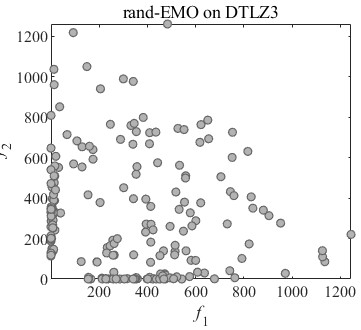}
                    \includegraphics[scale=0.27]{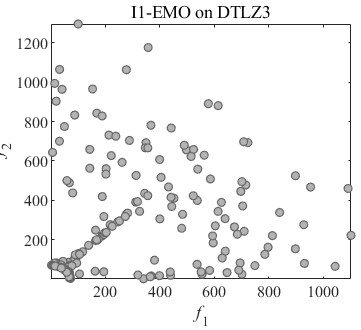}
                    \includegraphics[scale=0.27]{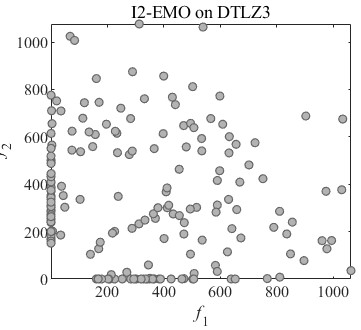}
                    \includegraphics[scale=0.27]{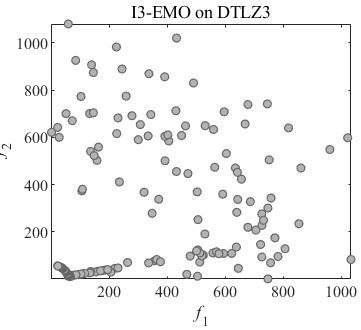}
                    \includegraphics[scale=0.27]{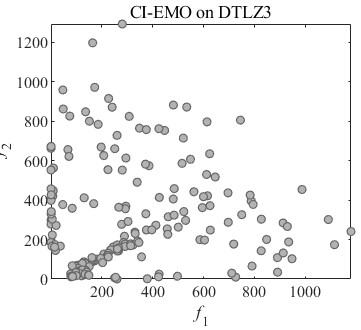}
                }
                
                \resizebox{\linewidth}{!}{
                    \includegraphics[scale=0.27]{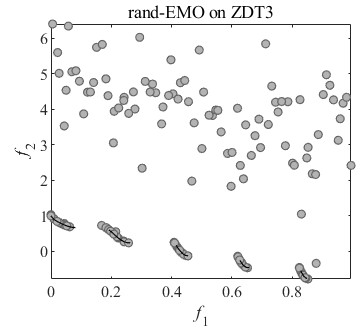}
                    \includegraphics[scale=0.27]{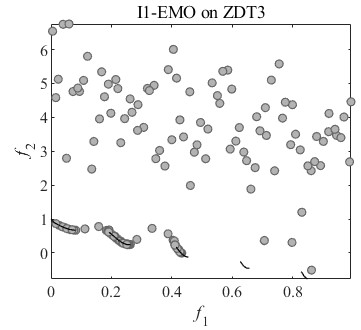}
                    \includegraphics[scale=0.27]{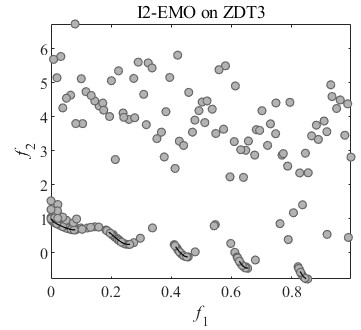}
                    \includegraphics[scale=0.27]{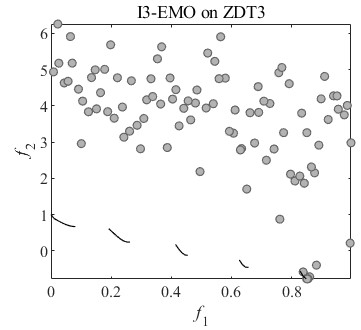}
                    \includegraphics[scale=0.27]{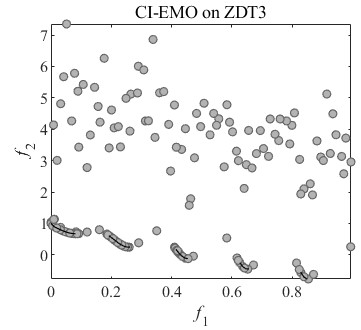}
    		}
    		\caption{Evaluated solutions obtained five algorithms under comparison on 2-objective DTLZ1-3 and ZDT3.}
    		\label{Figure_3_CI}
	\end{figure*}

        \begin{table}[h]\footnotesize
		  \centering
		  \caption{\centering IGD+ RESULTS (MEAN AND STANDARD DEVIATION) OBTAINED BY THE PROPOSED CI-EMO AND THE VARIANT WITH SAME WEIGHTS}
          \resizebox{0.7\linewidth}{!}{
          \begin{tabular}{|c|c|c|c|c|}
            \hline
            Problem & M     & D     & CI-EMO-SW & CI-EMO \\
            \hline
            \multirow{0.5}[4]{*}{DTLZ1} & 2     & 8     & \textbf{2.4563e+1 (8.93e+0) +} & 3.0662e+1 (8.78e+0) \\
            \cline{2-5}      & 3     & 6     & 9.8818e+0 (5.23e+0) $\approx$ & \textbf{9.4145e+0 (4.68e+0)} \\
            \hline
            \multirow{0.5}[4]{*}{DTLZ2} & 2     & 8     & 3.5816e-3 (8.43e-4) $\approx$ & \textbf{3.4063e-3 (4.00e-4)} \\
            \cline{2-5}      & 3     & 6     & \textbf{2.0913e-2 (8.40e-4) +} & 2.2150e-2 (1.83e-3) \\
            \hline
            \multirow{0.5}[4]{*}{DTLZ3} & 2     & 8     & \textbf{7.1682e+1 (3.56e+1) $\approx$} & 7.6125e+1 (2.95e+1) \\
            \cline{2-5}      & 3     & 6     & \textbf{3.1452e+1 (1.53e+1) $\approx$} & 4.0630e+1 (1.72e+1) \\
            \hline
            \multirow{0.5}[4]{*}{DTLZ4} & 2     & 8     & 1.9844e-1 (1.39e-1) $\approx$ & \textbf{1.9344e-1 (1.35e-1)} \\
            \cline{2-5}      & 3     & 6     & 1.0054e-1 (3.99e-2) $\approx$ & \textbf{9.2780e-2 (3.91e-2)} \\
            \hline
            \multirow{0.5}[4]{*}{DTLZ5} & 2     & 8     & 3.5905e-3 (8.66e-4) $\approx$ & \textbf{3.4790e-3 (4.10e-4)} \\
            \cline{2-5}      & 3     & 6     & 2.9777e-3 (1.99e-4) $\approx$ & \textbf{2.9033e-3 (1.68e-4)} \\
            \hline
            \multirow{0.5}[4]{*}{DTLZ6} & 2     & 8     & 1.1374e+0 (3.69e-1) - & \textbf{8.8361e-1 (3.18e-1)} \\
            \cline{2-5}      & 3     & 6     & 3.4928e-1 (1.41e-1) $\approx$ & \textbf{3.2192e-1 (1.67e-1)} \\
            \hline
            \multirow{0.5}[4]{*}{DTLZ7} & 2     & 8     & 2.0077e-3 (1.06e-4) $\approx$ & \textbf{1.9779e-3 (6.93e-5)} \\
            \cline{2-5}      & 3     & 6     & \textbf{2.2024e-2 (8.34e-4) +} & 2.2759e-2 (6.08e-4) \\
            \hline
            ZDT1  & 2     & 8     & 3.4690e-3 (5.67e-4) - & \textbf{2.9639e-3 (3.48e-4)} \\
            \hline
            ZDT2  & 2     & 8     & \textbf{2.6707e-3 (1.23e-4) $\approx$} & 2.6829e-3 (1.33e-4) \\
            \hline
            ZDT3  & 2     & 8     & 5.9452e-2 (2.76e-2) - & \textbf{2.6870e-3 (5.41e-4)} \\
            \hline
            ZDT4  & 2     & 8     & 2.6833e+1 (9.19e+0) $\approx$ & \textbf{2.2829e+1 (1.14e+1)} \\
            \hline
            ZDT6  & 2     & 8     & 4.9157e-1 (2.63e-1) $\approx$ & \textbf{4.1481e-1 (1.10e-1)} \\
            \hline
            \multicolumn{3}{|c|}{+/-/$\approx$} & 3/3/13 &  \\
            \hline
            \end{tabular}%
        }
        \label{tab:CI-EMO-SW}%
        \end{table}%

         \begin{figure*}[ht]
    		\centering
    		\resizebox{\linewidth}{!}{
    			\includegraphics[scale=0.27]{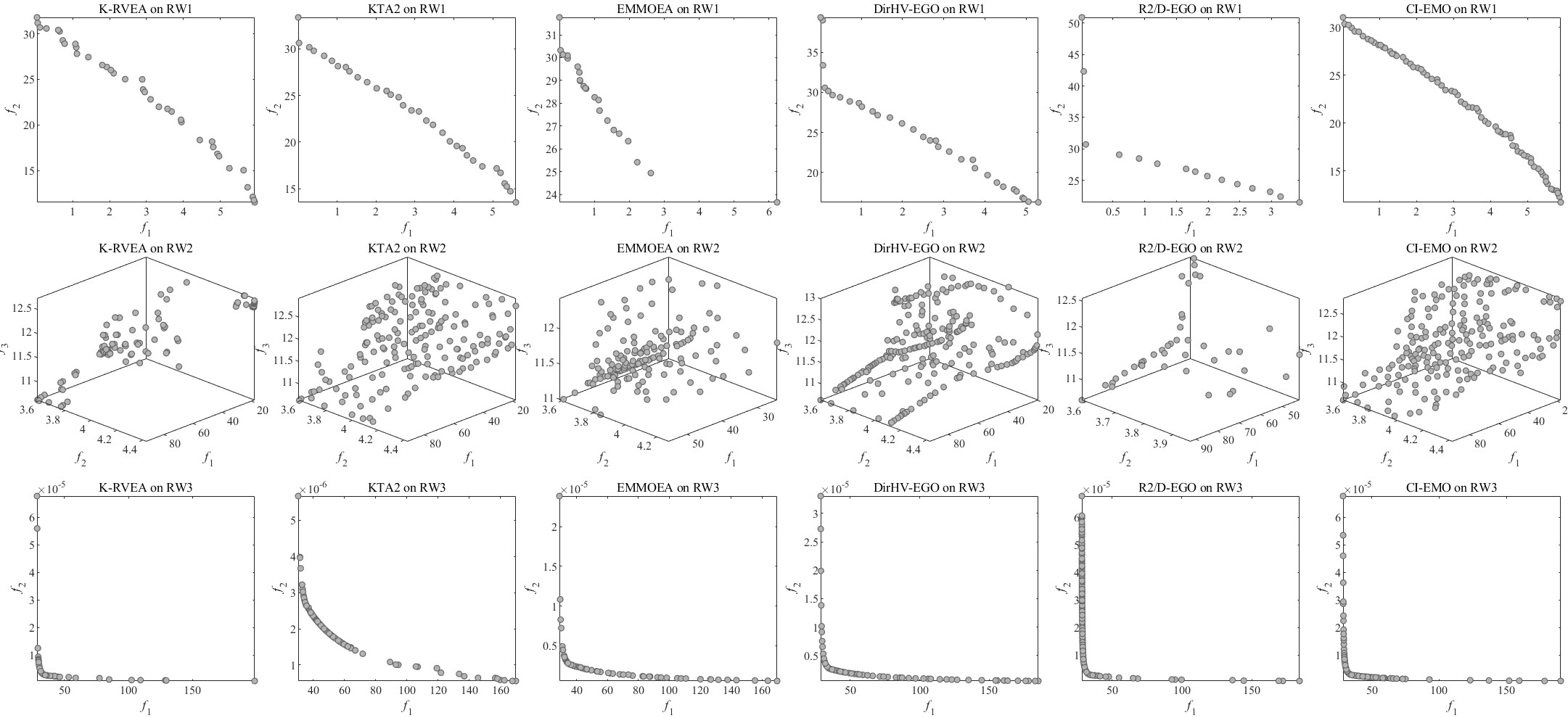}
    		}
    		\caption{The nondominated solutions obtained six algorithms under comparison on three real-world problems, gear train design problem (RW1), car side impact design problem (RW2), and two bar plane truss (RW3).}
    		\label{Figure_5_RW}
	\end{figure*}
	
\end{document}